\pdfoutput=1

\documentclass[11pt]{article}

\usepackage[final]{acl}

\usepackage{times}
\usepackage{latexsym}

\usepackage[T1]{fontenc}

\usepackage[utf8]{inputenc}

\usepackage{microtype}

\usepackage{inconsolata}

\usepackage{graphicx}

\usepackage{CJKutf8}

\usepackage{subcaption}

\usepackage{amsmath,amsfonts,bm}

\def\eqref#1{equation~\ref{#1}}

\def\1{\bm{1}}

\def\vb{{\bm{b}}}
\def\vc{{\bm{c}}}

\def\vp{{\bm{p}}}

\def\vs{{\bm{s}}}

\def\vz{{\bm{z}}}

\DeclareMathAlphabet{\mathsfit}{\encodingdefault}{\sfdefault}{m}{sl}
\SetMathAlphabet{\mathsfit}{bold}{\encodingdefault}{\sfdefault}{bx}{n}

\def\sS{{\mathbb{S}}}
\def\sT{{\mathbb{T}}}

\def\sV{{\mathbb{V}}}

\newcommand{\R}{\mathbb{R}}

\usepackage{amsthm}

\theoremstyle{plain}

\theoremstyle{definition}
\newtheorem{definition}{Definition}[section]

\theoremstyle{definition}

\theoremstyle{definition}

\theoremstyle{remark}

\theoremstyle{plain}

\usepackage{cleveref}
\usepackage{float}
\usepackage{enumitem}
\usepackage{placeins}

\newcommand{\jamie}[1]{}

\newcommand{\harsh}[1]{}
\newcommand{\ilia}[1]{}
\definecolor{marikascolor}{rgb}{0.0, 0.42, 0.24} 
\newcommand{\ms}[1]{}

\newcommand*{\escape}[1]{\texttt{\textbackslash#1}}
\definecolor{carminered}{rgb}{1.0, 0.0, 0.22}
\definecolor{ao}{rgb}{0.0, 0.5, 0.0}
\definecolor{applegreen}{rgb}{0.55, 0.71, 0.0}
\definecolor{emerald}{rgb}{0.31, 0.78, 0.47}
\definecolor{greenncs}{rgb}{0.0, 0.62, 0.42}
\definecolor{forestgreen}{rgb}{0.13, 0.55, 0.13}

\newcommand{\custompar}[1]{\vspace{1mm} \noindent{\bf #1}\:}

\definecolor{match}{rgb}{0.0706, 0.3725, 0.8588}
\definecolor{mismatch}{rgb}{0.8706, 0.1059, 0.0353}

\usepackage{hyperref}
\hypersetup{
    pdftitle={Measuring memorization in language models via probabilistic 
    extraction}
}

\title{Measuring memorization in language models via probabilistic extraction\looseness=-1}

\author{Jamie Hayes\thanks{Corresponding author: \texttt{jamhay@google.com}, \texttt{marikaswanberg@google.com}, \texttt{afedercooper@gmail.com}}\\
  Google DeepMind
  \And
  Marika Swanberg$^*$\\
  Google
  \And
  Harsh Chaudhari \\
  Google DeepMind \& Northeastern
  \AND
  Itay Yona \\
  Google DeepMind
  \And
  Ilia Shumailov \\
  Google DeepMind
  \And
  Milad Nasr \\
  Google DeepMind
  \AND
  Christopher A. Choquette-Choo \\
  Google DeepMind
  \And
  Katherine Lee \\
  Google DeepMind
  \And
  A. Feder Cooper$^*$ \\
  Microsoft Research \& Stanford
}

\begin{document}
\maketitle

\begin{abstract}
\vspace{-.1cm}
Large language models (LLMs) are susceptible to 
memorizing training data, raising concerns about the potential extraction of sensitive information at generation time. 
Discoverable extraction is the most common method for  measuring this issue: 
split a training example into a prefix and suffix, then prompt the LLM with the prefix, and deem the example extractable if the LLM generates the matching suffix using greedy sampling.
This definition yields a yes-or-no determination of whether extraction was successful with respect to a \emph{single} query. 
Though efficient to compute, we show that this definition is unreliable because it does not account for non-determinism present in more realistic (non-greedy) sampling schemes, for which LLMs  produce a range of outputs for the same prompt.
We introduce \emph{probabilistic} discoverable extraction, which, without additional cost,  
relaxes discoverable extraction by considering \emph{multiple} queries to quantify the probability of extracting a target sequence. 
We evaluate our probabilistic measure across different models, sampling schemes, and training-data repetitions, and find that 
this measure provides more nuanced information about extraction risk compared to traditional discoverable extraction.\looseness=-1 
\end{abstract}
\vspace{-.3cm}
\section{Introduction}\label{sec:intro}
Large language models (LLMs) are susceptible to 
memorizing 
pieces of their training data, raising concerns about the potential extraction of sensitive information in the training dataset at generation time~\citep{carlini2021extracting, carlini2022quantifying, shi2023detecting, zhang2023counterfactual, smith2023identifying,  lee2023talkin, biderman2024emergent, duan2024uncovering, cooper2024files, huang2024demystifying,  bordt2024elephants}.\footnote{{\footnotesize{This paper covers a very restricted definition of ``memorization'': 
    whether a generative model can be induced to generate near-facsimiles of some training examples when prompted with appropriate instructions. 
    Models do not ``contain'' bit-wise or code-wise copies of their training data.
    Rather, 
    If a model can be induced to generate very close copies of certain training 
    examples by supplying 
    such instructions to guide the model's statistical generation process, then that model is said to have ``memorized'' those examples.
    This is an area of active 
    ongoing research.}\looseness=-1}
} 
This issue has attracted significant attention, leading researchers to routinely report training-data extraction rates in technical reports introducing new LLMs~\citep{chowdhery2023palm,anil2023palm,reid2024gemini,team2024gemma,team2024gemma2,biderman2023pythia, grattafiori2024llama3herdmodels}.
There are numerous ways to measure these rates, but one of the most common is to quantify \textbf{discoverable extraction}: 
split an LLM's training example into a prefix and suffix, prompt the LLM with the prefix, and deem the example extractable if the LLM generates the sequence that matches the suffix~\citep{carlini2021extracting,carlini2022quantifying,nasr2023scalable} (Section~\ref{sec:prelim}).\looseness=-1

Discoverable extraction is simple and efficient to compute; 
however, it has drawbacks that may make it an unreliable estimate for a model's true extraction rate.
Notably, the definition for discoverable extraction does not account for the  non-deterministic nature of LLMs.
It yields a yes-or-no determination of whether extraction was successful with respect to a \emph{single} user query, most typically executed with deterministic greedy sampling. 
But, of course, in realistic production settings, users  may query the model multiple times and with 
non-deterministic sampling schemes, where  
multiple queries with the same prompt  can  result  in a range of different  generations. 
So, by only generating a single sequence to check for a match with the target, discoverable extraction may miss cases where a match could have been found if more than one sequence had been generated. 
Further, extracting sensitive training data  
even once out of multiple queries could be problematic, as an adversary (e.g., a hacker checking credit card numbers) may have external means of verifying the sensitive information's validity. 
It is therefore reasonable to quantify the number of sequences that need to be generated before a particular target example becomes extractable.\looseness=-1 

Following from this motivation, we introduce \emph{probabilistic} discoverable extraction: 
a relaxation of discoverable extraction that considers \emph{multiple} queries in order to quantify the probability of extracting a particular target sequence. 
That is, our new definition for  \textbf{\bm{$(n,p)$}-discoverable extraction} 
quantifies the number of attempts $n$ an adversary would need to make to extract a target sequence at least once with 
probability $p$ under a given sampling scheme  
(Section~\ref{sec:def}). 
We benchmark $(n,p)$-dis\-cov\-er\-able extraction rates for different sampling schemes, settings of $n$ and $p$, model families and sizes, and repetitions of target data  (Section~\ref{sec:experiments}). 
In summary, $(n,p)$-discoverable extraction\looseness=-1 
\begin{itemize}[leftmargin=.5cm,topsep=.1cm,itemsep=0cm]
    \item Provides more meaningful  measurements of extraction rates than greedy-sampled discoverable extraction, which, by comparison,  
    underestimates extraction rates 
    even for modest values of $n$ and $p$ (Sections~\ref{sec:def:def} \&~\ref{ssec: n_and_p}).
    The gap between the two rates increases for larger models and more repetitions of the target  (\Cref{ssec: gap}).\looseness=-1
    
    \item Captures the risk that a particular target can be extracted under a given non-deterministic sampling scheme (Sections~\ref{sec:def:def} \&~\ref{ssec:relative_risk}).\looseness=-1  
    
    \item Can be approximated with high-fidelity using 
    just one query---i.e., \emph{with no overhead compared to discoverable extraction} (Section~\ref{sec:def:hps}). 
    \item Is easily extended to quantify  
    extraction 
    for 
    non-verbatim matches to the target  
    (Section~\ref{sec:def:apprx}).\looseness=-1 

\end{itemize}

\vspace{-.1cm}
\section{Preliminaries and related work}\label{sec:prelim}
\vspace{-.05cm}

We begin with prior work on discoverable extraction (Section~\ref{sec:prelim:discoverable}) and relevant background on different 
sampling schemes (Section~\ref{sec:prelim:sampling}).\looseness=-1

\subsection{Discoverable extraction}\label{sec:prelim:discoverable}
There are many different definitions for extraction in the literature (Appendix~\ref{app:sec:othermem}), but one of the most popular is \textbf{discoverable extraction}~\citep{anil2023palm,reid2024gemini,team2024gemma,team2024gemma2,kudugunta2024madlad,grattafiori2024llama3herdmodels, biderman2023pythia, kassem2024alpaca}. 
We adapt existing definitions for  discoverable extraction~\citep{carlini2021extracting, carlini2022quantifying, nasr2023scalable}, described in terms of an arbitrary training example $\vz$, model $f_\theta$, and sampling scheme $g_\phi$.\looseness=-1 
 
For a $j$-length sequence of tokens $\vz\! =\! (z_1, \ldots, z_j)$ and indices $1\! \leq\! h \!\leq\! i\!\leq\! j$, we use $\vz_{h:i}$ to denote tokens $z_h, \ldots, z_i$. 
Let $f_{\theta}\!: \sV^j \rightarrow \mathcal{P}(\sV)$ be a $\theta$-parameterized LLM that takes a sequence of $j$ tokens from a vocabulary $\sV$ and outputs a probability distribution $\mathcal{P}(\sV)$ over $\sV$. 
Let $g_{\phi}\!: \mathcal{P}(\sV) \rightarrow \sV$ be a sampling scheme parameterized by scheme-specific hyperparameters $\phi$, which takes as arguments a probability distribution over vocabulary $\sV$ and selects a token from $\sV$. 
Finally, for some initial sequence $\vz$, let $(g_{\phi} \circ f_\theta)^k(\vz)$ denote the autoregressive process of repeatedly generating a distribution over the token vocabulary, sampling a token from this distribution, and adding the token to the sequence $k>0$ times, starting from the initial sequence $\vz$.
We will use \textbf{query} to denote this entire autoregressive process of generating multiple (up to size $k$) tokens. \looseness=-1

\begin{definition}[\textbf{Discoverable extraction}]\label{def:discoverable_extraction}
Given a training example $\vz$ that is split into an $a$-length prefix $\vz_{1:a}$ and a $k$-length suffix $\vz_{a+1:a+k}$, $\vz$ is \emph{discoverably extractable} if $(g_{\phi} \circ f_\theta)^k(\vz_{1:a})=\vz_{1:a+k}$.
\end{definition}

\noindent In other words, $\vz = \vz_{1:a} \parallel \vz_{a+1:a+k}$;
we use the first $a$ tokens of a training example $\vz$ as the input prompt to the generation process, and then check if the sequence generated under the composition of model $f_{\theta}$ and sampling scheme $g_{\phi}$ matches verbatim the remaining $k$ tokens in the example. 
We refer to this as \textbf{one-shot extraction}, given that it returns a binary, yes-or-no determination for a single query. 
We will relax this when we revisit extraction from a probabilistic perspective  in Section~\ref{sec:def}.\looseness=-1 

To measure discoverable extraction in practice, prior work has relied on  different instantiations of Definition~\ref{def:discoverable_extraction}. 
This includes varying both the length of prefix prompts and the minimum-length generated suffix that qualifies as extraction.
For example, \citet{carlini2022quantifying} test prefix lengths with prompts ranging from $50$ to $500$ tokens, and consider a training example to be extracted if the model generates the subsequent $50$ tokens in the example.
Alternatively, \citet{biderman2023pythia} set the prefix and suffix size to $32$ tokens.\looseness=-1 

Further, the sequence a model $f_\theta$ generates is entirely dependent on the choice of sampling scheme $g_\phi$ that defines how a token is selected from the output distribution over the model's vocabulary $\mathcal{P}(\sV)$. 
In most prior work, the common choice 
for $g_\phi$ is \textbf{greedy sampling}, which generates a sequence by selecting the highest-probability token, conditioned on the previous tokens, at each step. 
Focusing on one-shot extraction is justified in this setting, given that the output is deterministic.  
Work by \citet{carlini2022quantifying} is an  exception; while they also focus on greedy sampling, they additionally analyze one-shot discoverable extraction with beam search.\looseness=-1

\subsection{Alternate sampling schemes}\label{sec:prelim:sampling}  
While greedy sampling is the most common choice for $g_\phi$ in prior work on discoverable extraction, it provides an incomplete picture. 
Many production language models are deployed with non-greedy schemes, 
and API users are often free to decide which 
scheme to use. 
Other sampling schemes, e.g., \citet{fan2018hierarchical, basu2020mirostat, vijayakumar2016diverse}, are  desirable for increasing output diversity. 
Further, although greedy sampling locally selects the most likely \emph{next token} it may not select the overall most likely \emph{sequence}, which can decrease generation quality.  
Below, we briefly discuss popular choices for $g_\phi$.\looseness=-1  

\custompar{Random sampling with temperature.} 
Given a sequence of $t-1$ tokens $\vz_{1:t-1}$, we sample 
\vspace{-.15cm}
\begin{align}
\label{eq:temp}
\textstyle
\mathcal{P}(z_t \mid \vz_{1:t-1}) = \frac{\mathrm{e}^{\frac{y(z_t)}{T}}}{\sum_{v\in\sV} \mathrm{e}^{\frac{y(v)}{T}}},
\end{align}
\vspace{-.25cm}

\noindent where $y(v)$ denotes the logit value of token $v\in\sV$, and $T\in\R_{>0}$ is a \textbf{temperature} value that controls the flatness (or sharpness) of the probability distribution $\mathcal{P}(\sV)$. 
$T=1$ samples from the base distribution output by the model $f_{\theta}$,  larger $T$ increases  diversity, and, as $T\rightarrow 0$, Equation (\ref{eq:temp}) converges to deterministic greedy sampling. 

\custompar{Top-$k$ sampling.}
In $\mathcal{P}(z_t | \vz_{1:t-1}$), all but the $k$ tokens with the highest probabilities have their probabilities set to 0, the non-zero probabilities are  normalized, and the next token is sampled accordingly. 
(If $k=1$, this is identical to greedy sampling.)

\citet{carlini2022quantifying} argue  that using sampling schemes like these, which have higher degrees of associated randomness compared to deterministic  greedy sampling, are ``antithetical'' to maximizing discoverable extraction; 
such schemes ``maximi[ze] linguistic novelty'' and so are less likely to generate outputs that match existing text. 
Nonetheless, these schemes may be advantageous for more reliably identifying extraction in realistic settings where, to explore different outputs, a  user may query the model \emph{multiple} times with the same prompt. 

\section{Probabilistic discoverable extraction}\label{sec:def}

The more realistic setting of non-deterministic  sampling and multiple queries motivates measuring extraction from a probabilistic perspective.   
Rather than taking a one-shot approach, it is reasonable to  
quantify the number of sequences that need to be generated before a target example becomes likely to be  extracted under a chosen sampling scheme. 

Following from this motivation, we introduce 
\textbf{$\bm{(n,p)}$-discoverable extraction}, which 
captures the capabilities of a regular user 
that can query the model $n$ times 
to extract verbatim a target sequence with probability $p$ (Section~\ref{sec:def:def}).
Under non-deterministic sampling schemes, many extractable targets  are unlikely to be guaranteed to be generated (i.e., $p\!\!=\!\!1$) with one query, but  
many targets may be generated at least once given enough attempts ($n\!>\!1$). 
Thus, our definition captures a continuous notion of the risk of extracting the target example;  
it  
provides a more meaningful estimate of extraction risk than can be gleaned from one-shot, yes-or-no tests (Section~\ref{sec:prelim:discoverable}).\looseness=-1 

We then discuss how  $n$ and $p$ are directly related through a simple equation 
such that, in practice, it is possible to measure a high-fidelity approximation of $(n,p)$-discoverable extraction with just one query---\emph{with no overhead compared to traditional discoverable extraction} (Section~\ref{sec:def:hps}). 
We then briefly describe a variant of $(n,p)$-discoverable extraction that measures the non-verbatim extraction of target sequences (Section~\ref{sec:def:apprx}),  
and draw connections between our contributions and other work (Section~\ref{sec:def:connections}).\looseness=-1

\begin{figure*}[t!]
    \centering
    \begin{subfigure}{0.54\textwidth}
        \centering
        \includegraphics[width=.98\linewidth]{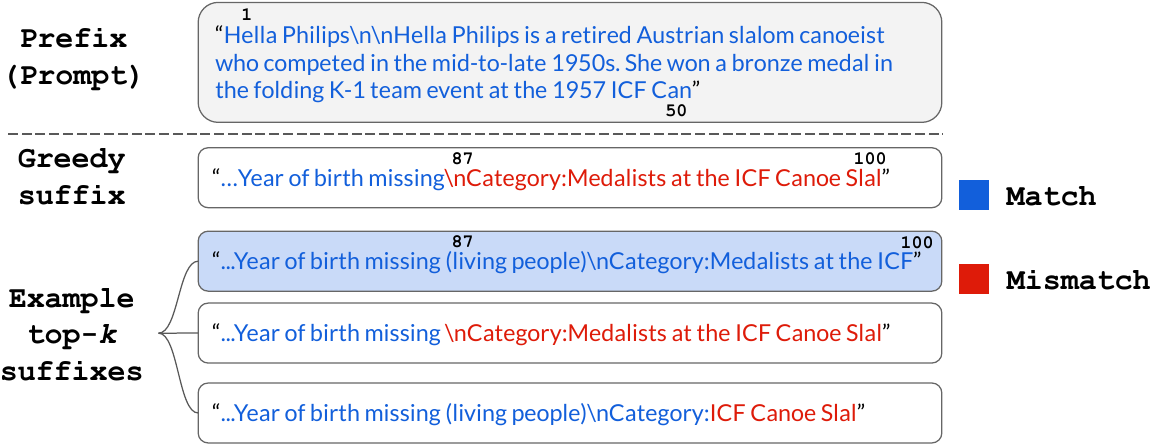}
        \label{fig:greedy_fail:tokens}
    \end{subfigure}
    \hfill
    \begin{subfigure}{0.45\textwidth}
        \centering
        \includegraphics[width=0.85\linewidth]{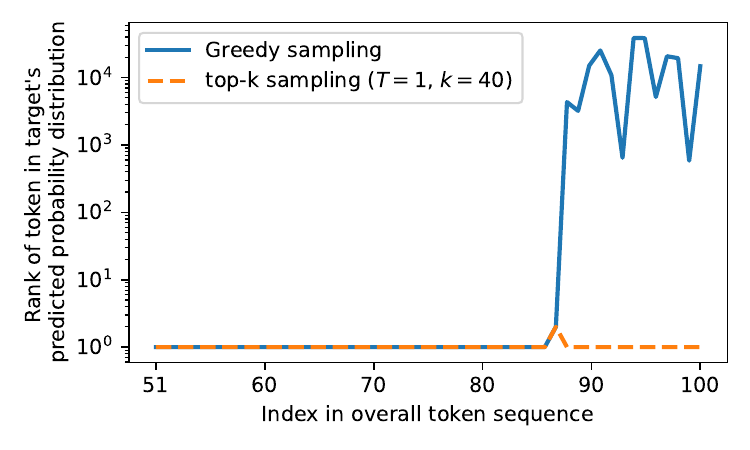}
        \label{fig:greedy_fail:plot}
    \end{subfigure}
    \vspace{-.3cm}
    \caption{\textbf{Left:} The prefix $\vz^t_{1:50}$, and portions of the greedy-sampled suffix and of  example top-$k$-sampled suffixes for Pythia 12B. 
    \textcolor{match}{Blue} indicates a match with the target, \textcolor{mismatch}{red} a mismatch. 
    \textbf{Right}: 
    For each successive token that is decoded by greedy and top-$k$ ($k\!=\!40$, $T\!=\!1$) sampling, we plot the probability rank with respect to the target suffix token.  
    At index $87$, the target token has rank $2$; greedy sampling does not select this token, after which the greedy-generated sequence diverges from the target. 
    In contrast, top-$k$ sampling picks the rank-$2$ token and proceeds to extract the target sequence correctly (with probability $16.2\%$). 
     Note that, if greedy sampling had selected the rank-$2$ token at index $87$, then it would have generated the target, as the remaining target tokens all have rank-$1$.\looseness=-1
     }
 \label{fig:greedy_fail}
 \vspace{-.35cm}
\end{figure*}

\subsection{Defining $(n,p)$-discoverable extraction}\label{sec:def:def}

We define our probabilistic relaxation of discoverable extraction (Definition~\ref{def:discoverable_extraction}) with two new hyperparameters: 
number of queries $n$ and probability $p$:\looseness=-1 

\begin{definition}[\textbf{$(n, p)$-discoverable extraction}] 
\label{def:np_discoverable_extraction}
Given a training example $\vz$ that is split into an $a$-length prefix $\vz_{1:a}$ and a $k$-length suffix $\vz_{a+1:a+k}$, $\vz$ is \emph{$(n, p)$-discoverably extractable} 
if 
\vspace{-.2cm}
\begin{align*}
    \Pr\Big(\cup_{w\in[n]} (g_{\phi} \circ f_\theta)_w^k(\vz_{1:a})=\vz_{1:a+k}\Big)\geq p,
\end{align*}
\vspace{-.4cm}

\noindent where  $(g_{\phi} \circ f_\theta)_w^k(\vz_{1:a})$ represents the $w$-th (of $n$)
independent execution of the autoregressive process of 
generating a distribution over the token vocabulary, sampling a token from this distribution, 
and adding the token to the sequence $k > 0$ times, starting from the same initial sequence $\vz_{1:a}$.
\end{definition}

That is, in total, we generate $n$ independent sequences  by sampling the output distribution of model $f_\theta$ using scheme $g_\phi$. 
If the probability of generating $\vz_{a+1:a+k}$ \emph{at least once} is larger than $p$, then we say $\vz$ is $(n, p)$-discoverably extractable with respect to $f_\theta$ and $g_\phi$. 
When the sampling scheme and model are clear from context, we simply say the target is $(n,p)$-discoverably extractable.
One can also view $(n, p)$-discoverable extraction as the probability that, together, $f_\theta$ and $g_\phi$ provide an $n$-sized anonymity set: a set of $n$ data points within which the true training example can hide. 
Note that we can write discoverable extraction in terms of this probabilistic relaxation by setting $n=p=1$. 
Also note that $n$ and $p$ are directly related. 
We will revisit this observation in relation to choosing $n$ and $p$ in practice (Section~\ref{sec:def:hps}).\looseness=-1

Next, we describe 
how 
$(n,p)$-discoverable extraction offers several advantages over standard discoverable extraction instantiated with greedy sampling (\Cref{def:discoverable_extraction}). 
Through the example in Figure~\ref{fig:greedy_fail}, we show how greedy sampling can underestimate extraction. 
As a result, research and model-release reports that rely on greedy-sampled discoverable extraction~\citep[e.g.,][]{reid2024gemini,grattafiori2024llama3herdmodels} could substantially differ from the amount of extraction seen by an end user. 
We then briefly discuss how multiple queries can be used to  
capture extraction risk.\looseness=-1 

\custompar{Greedy sampling misses instances of extraction.} 
We illustrate this failure mode for greedy-sampled discoverable extraction for target prefix $\vz^t_{1:50}$ with an example  (\Cref{fig:greedy_fail}). 
The target suffix $\vz^t_{51:100}$ has a higher overall  likelihood of being generated, compared to the greedy-sampled suffix $\vz^g_{51:100}$. 
In always selecting the locally-most-likely next token (Section~\ref{sec:prelim:sampling}), 
greedy sampling 
generates an output that does not match the overall-more-likely target;\footnote{The normalized edit distance between greedy and target outputs is $13.6\%$ in character space and $16\%$ in token space.\looseness=-1}  
greedy sampling fails to extract the target.\looseness=-1

Figure~\ref{fig:greedy_fail} explores how this happens in more detail. 
For each token $i$ in the target suffix $\vz^t_{51:100}$, 
we compute the probabilities for each possible next token in the vocabulary (i.e., $\mathcal{P}(z^t_{i} | \vz^t_{1:i-1})$) and we rank them, with rank $1$ reflecting the highest-probability next token. 
For the first 36 suffix tokens (through overall index $86$), the rank-$1$ token 
aligns with the actual target token. 
So, greedy sampling, which always locally selects the rank-$1$ token, selects these 
tokens 
(i.e, $\vz^g_{51:86} = \vz^t_{51:86}$). 
But, for the next token, the target token is the \emph{second-highest probability}, rank-$2$ token, so greedy sampling does not select it (i.e., $z^g_{87} \ne z^t_{87}$). 
With this deviation, the rest of the greedy-sampled sequence continues to diverge from the target sequence; in each iteration, the locally highest-probability token 
differs (often significantly in rank) from the (overall-higher-probability) 
target token.\looseness=-1 

In contrast to greedy sampling, using a probabilistic sampling scheme allows for the possibility of selecting the rank-$2$ token at index $86$. 
At this point,  
it becomes highly probable that the entire target suffix is extracted. 
We demonstrate this with top-$k$ sampling ($T=1$, $k=40$), where the target has probability of $16.2\%$ of being extracted in one shot. 
Additional examples are in Appendix~\ref{app: more_why}. 

\custompar{Extraction risk.} 
One can view $(n, p)$-dis\-cov\-er\-a\-ble extraction as capturing the risk of a user extracting a particular target sequence 
as a function of the number of queries. 
This setting is reflective of 
production users:  
users can (and do) query LLMs many times. 
In 
Figure~\ref{fig:greedy_fail}, a user would only need $6$ queries (in expectation) before generating the target, since it occurs with probability $16.2\%$. 
A user who extracts sensitive information after several attempts  could be just as successful at exploiting this information, compared to if they had extracted it in one shot. 
This is because many sensitive sequences (e.g., phone numbers, credit card numbers) can be verified through external means. 

\subsection{How should one set on $n$ and $p$?}\label{sec:def:hps}
Our definition introduces two related hyperparameters: number of queries $n$ and probability $p$. 
In general, if it is easy to extract a particular target example $\vz$, the number of queries $n$ to generate $\vz$ at least once is small (and thus $p$ is large). 
The reverse holds for targets that are challenging to extract. 
That is, 
a specific model, sampling scheme, and example $\vz$ 
define a trade-off between $n$ and $p$ such that $\vz$ is $(n,p)$-discoverably extractable. 
Because of this trade-off, 
we can approximate $(n,p)$-discoverable extraction with just one query---with no overhead compared to 
discoverable extraction (Definition~\ref{def:discoverable_extraction}).\looseness=-1 

To see this, let us say that  $p_\vz$ is the probability of generating a suffix $\vz_{a+1:a+k}$ for prefix $\vz_{1:a}$, given a  model, sampling scheme, and example $\vz = \vz_{1:a}\!\!\parallel\!\!\vz_{a+1:a+k}$.  
This means that  
the probability of \emph{not} generating $\vz_{a+1:a+k}$ in a single draw from the sampling scheme is $1-p_\vz$, and the probability of not generating $\vz_{a+1:a+k}$ in $n$ independent draws 
is $(1-p_\vz)^n$. 
Therefore, example $\vz$ is $(n,p)$-discoverably extractable for $n$ and $p$ that satisfy $1-(1-p_\vz)^n \geq p$.
Equivalently,\looseness=-1 
\vspace{-.1cm}
\begin{equation}\label{eq:npmem}
    n \geq \frac{\log(1-p)}{\log(1-p_\vz)}.
\end{equation}
\vspace{-.3cm}

\noindent In other words, we can easily find $n$ given a fixed $p$ and the probability of generating a sequence $p_\vz$, and vice versa. 
In practice, for a given  
$\vz$, we can compute $p_\vz$ with just one query (Appendix~\ref{app:sec:setup:computing}). 
This lets us use Equation (\ref{eq:npmem}), where we fix a desired minimum extraction probability $p$ and find the corresponding queries $n$. 
In expectation, $ n=\lceil\frac{1}{p_\vz}\rceil$.\looseness=-1

\begin{figure}[t]
  \begin{center}
\includegraphics[width=\linewidth, height=1.35in]{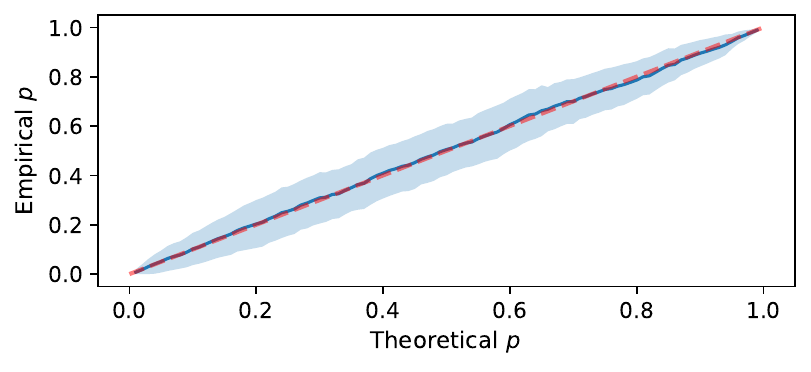}
  \end{center}
  \vspace{-.35cm}
  \caption{For $250$ examples in the Pile (Wikipedia subset) and Pythia 6.9B, we check that generating 
  $n\!\!=\!\!1000$ sequences and computing the probability a training example appears at least once in the set (\textbf{empirical $p$}) matches the \textbf{theoretical $p$} 
  using Equation (\ref{eq:npmem}).\looseness=-1
  }
  \label{fig:emp_p}
  \vspace{-.35cm}
\end{figure}

We verify that this one-query procedure for approximating $n$ and $p$ gives a reasonable estimate of $(n,p)$-discoverable extraction.
That is, we confirm that we can use one query to produce $p_\vz$ and Equation (\ref{eq:npmem}) for our measurements, rather than the more costly procedure of directly sampling a set of $n$ sequences to estimate $\hat{p}_\vz$ (the fraction of $n$ queries that generate the target suffix) and then computing $p$ (Appendix~\ref{app:sec:setup:computing}). 
In Figure~\ref{fig:emp_p}, we compare the two procedures.   
We plot the more costly, $n$-query-computed $p$ (i.e., empirical $p$) as a function of the corresponding $p$ computed using Equation (\ref{eq:npmem}) (i.e., theoretical $p$), and indeed the two match. 
For efficiency, we use Equation (\ref{eq:npmem}) in our experiments for verbatim extraction  (Section~\ref{sec:experiments}).\looseness=-1 

\custompar{Choosing an extraction tolerance.}  
The above are general observations about how the math works out; 
they do not reveal how one ought to set a ``reasonable'' threshold for  
the risk of extraction  
in practice.
What values for $n$ and $p$ would connote that a particular target is ``reasonably'' extractable? \looseness=-1 

We deliberately do not prescribe specific choices for $n$ and $p$, as ``reasonable'' choices depend on the type of information one is trying to extract~\citep{cooper2024files,cooper2024unlearning}. 
For example, one may be willing to release a model if $n$ is small and $p$ is high for extracting generic phrases, 
but this may not be tolerable for PII. 
We view this flexibility of $(n,p)$-discoverable extraction as a benefit: 
it allows practitioners to make fine-grained, context-specific decisions 
according to their respective levels of extraction-risk tolerance.\looseness=-1 

One possible way to reason about this tolerance is to set a threshold in terms of 
the expected computation limits of an adversary. 
If we assume that an adversary has a limited computation budget with which to query the model, we only need to ensure that $n$ is larger than this budget, 
in order to minimize the risk  of extraction. 
In practice, this $n$ could be enforced through rate limiting.\looseness=-1

\subsection{Extending to non-verbatim  extraction}\label{sec:def:apprx}

So far, we have considered probabilistic extraction of sequences that exactly match  the target. 
We now show how to relax  $(n,p)$-discoverable extraction  (Definition~\ref{def:np_discoverable_extraction}) to apply to non-verbatim matches.

\begin{definition}[\textbf{$(\epsilon, n, p)$-dis\-cov\-er\-able extraction}] 
\label{def:enp_discoverable_extraction}
For $\epsilon \in \R_{>0}$ and $\vb, \vc \in \sV^k$,   
define the set of $k$-length sequences 
    $\sS_\epsilon(\vb) = \{\vc \mid\mathsf{{dist}}(\vb, \vc) \leq \epsilon\}$, 
where $\mathsf{{dist}}\!\!: \sV^k \times \sV^k \rightarrow \R_{\geq 0}$ is a function that computes the distance between two $k$-length token sequences. 
Given a training example $\vz$ that is split into an $a$-length prefix $\vz_{1:a}$ and a $k$-length suffix $\vz_{a+1:a+k}$, $\vz$ is \emph{$(\epsilon, n, p)$-discoverably extractable} 
if 
\vspace{-.1cm}
\begin{align*}
    \Pr\Big(\cup_{w\in[n]} (g_{\phi} \circ f_\theta)_w^k(\vz_{1:a}) \in  \sT_\epsilon\Big) \geq p,
\end{align*}
\vspace{-.5cm}

\noindent where $\sT_\epsilon = \{\vz_{1:a} \parallel \vs \mid \vs \in   \sS_\epsilon(\vz_{a+1:a+k}) \}$. 
\end{definition} 

As in 
$(n,p)$-discoverable extraction, 
we generate $n$ independent sequences by sampling the output distribution of model $f_\theta$ using scheme $g_\phi$. 
But here, if the probability of generating \emph{any} suffix in $\sS_\epsilon(\vz_{a+1:a+k})$ \emph{at least once} is larger than $p$, then we say $\vz$ is $(\epsilon, n, p)$-discoverably extractable with respect to $f_\theta$ and $g_\phi$. 
In practice, this can be quite expensive to compute directly, as even for small $\epsilon$ the number of suffixes in $\sS_\epsilon(\vz_{a+1:a+k})$  may be very large. 
We provide additional discussion, including more efficient approximations, 
in Appendix~\ref{app:sec:nonverbatim}.\looseness=-1 

\subsection{Connections to other definitions}\label{sec:def:connections}

Our definition for $(n,p)$-discoverable extraction can be related to other work.  
Notably, 
\citet{carlini2019secret} motivate a measure of \textbf{canary memorization}  using rank perplexity, where a canary is a unique sequence deliberately inserted into the model's training data. 
This work considers an adversary that sequentially queries the model to extract  the canary, with guess/candidate canaries sorted from lowest to highest perplexity;  
the rank of the true canary quantifies how many guesses such an adversary would need to make before extracting it. 
While the secret-sharer attack that \citet{carlini2019secret} propose only involves one guess, it is natural to consider how extraction rates scale with multiple attempts---just as we do with $n$ queries  (Section~\ref{sec:experiments}).\looseness=-1  

Work by \citet{tiwari2025sequencelevelleakagerisktraining}, published shortly after ours, also studies extraction probabilities for different sampling strategies and models.  
They also consider non-verbatim probabilistic extraction (Section~\ref{sec:def:apprx}); 
we refer to Appendix~\ref{app:sec:nonverbatim} and their experimental results on this topic, and focus on verbatim memorization below. 
In Appendix~\ref{app:sec:othermem}, we provide extensive discussion on connections to canary memorization and other related definitions.\looseness=-1 

\section{Experiments}\label{sec:experiments}

We now demonstrate the utility of taking a probabilistic perspective on measuring extraction through experiments involving  different model families. 
We show how $(n,p)$-discoverable extraction rates change as a function of $n$ and $p$ (Section~\ref{ssec: n_and_p}),  and evaluate how these rates increase for larger model sizes and training-data repetitions (Section~\ref{ssec: gap}). 
We also verify that our analysis reflects valid estimates of training-data extraction.  
To do so, 
we compare the $(n,p)$-discoverable extraction rate to the corresponding rate of generating test-data examples   
(\Cref{ssec:relative_risk}). 
Altogether, our analysis illustrates that $(n,p)$-discoverable extraction provides more reliable measurements of extraction rates than greedy-sampled discoverable extraction, and also reveals a more nuanced picture of extraction risk in LLMs.\looseness=-1 

\custompar{Setup.} 
In this section, we analyze the  Pythia  model family~\citep{biderman2023pythia} and GPT-Neo 1.3B~\citep{gptneo}. 
In each experiment, we measure extraction rates with respect to $10,000$ examples drawn from the Enron dataset, which is contained in the Pile~\citep{gao2020pile}---the training dataset for both Pythia and GPT-Neo 1.3B.  
These extraction rates  
do not necessarily reflect the overall rates that would result for a representative sample of the entire training dataset. 
Throughout, we use one query and Equation (\ref{eq:npmem}) to compute $(n,p)$-discovable extraction (Definition~\ref{def:np_discoverable_extraction}). 
We use top-$k$ sampling ($k\!=\!40$,  
$T\!=\!1$) and compare to the one-shot, greedy-sampled discoverable extraction rate (\Cref{def:discoverable_extraction}).
Following \citet{carlini2021extracting}, in all cases, we use the first $50$ tokens of each example as the prefix, and the subsequent $50$ tokens as the suffix. 
In the Appendix, we provide results  on the OPT~\citep{zhang2022optopenpretrainedtransformer} and Llama model families~\citep{touvron2023llamaopenefficientfoundation}, as well as with different sampling schemes and training-data subsets.\looseness=-1  

\begin{figure*}[t]
\captionsetup[subfigure]{justification=centering}
  \centering
\begin{subfigure}[t]{0.49\textwidth}
\centering
    \includegraphics[width=\linewidth]{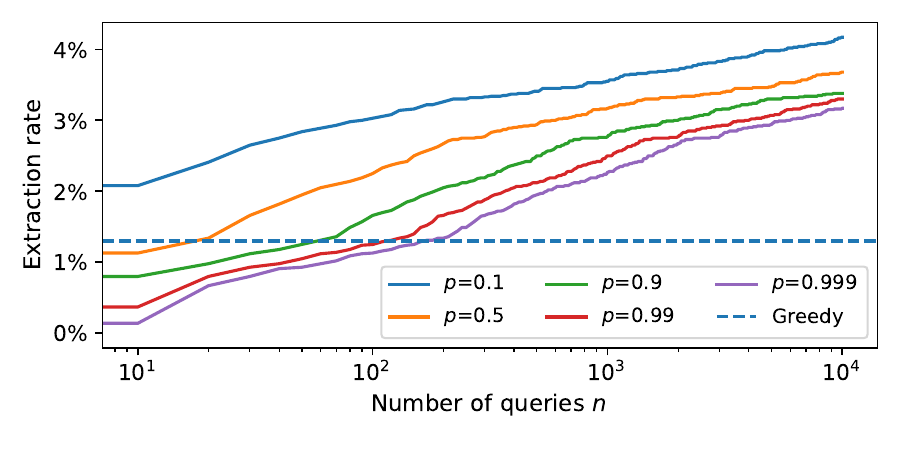}
        \vspace{-.6cm}
        \caption{Pythia 2.8B}
        \label{fig: topk2b}
\end{subfigure}
\hfill
\begin{subfigure}[t]{0.49\textwidth}
\centering
    \includegraphics[width=\linewidth]{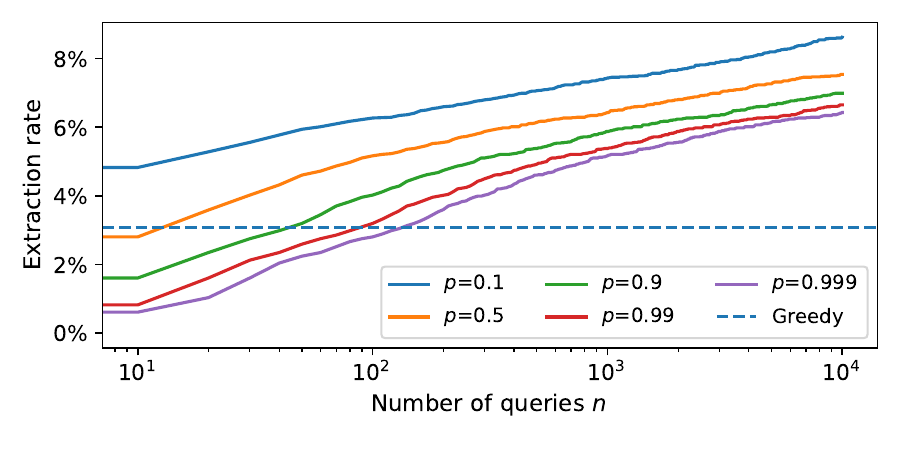}
        \vspace{-.6cm}
        \caption{Pythia 12B}
        \label{fig: topk12b}
\end{subfigure}
\vspace{-.2cm}
\caption{For $10,000$ examples from the Enron dataset, we plot variations in $(n, p)$-discoverable extraction rates for models of different sizes, according to different query budgets $n$ and minimum extraction probability $p$.\looseness=-1 
}
\label{fig: model_sizes}
\vspace{-.4cm}
\end{figure*}

\subsection{How extraction rates vary with $n$ and $p$}\label{ssec: n_and_p}

\begin{figure}[t!]
\vspace{-.2cm}
\centering
  \begin{center}
\includegraphics[width=.95\linewidth]{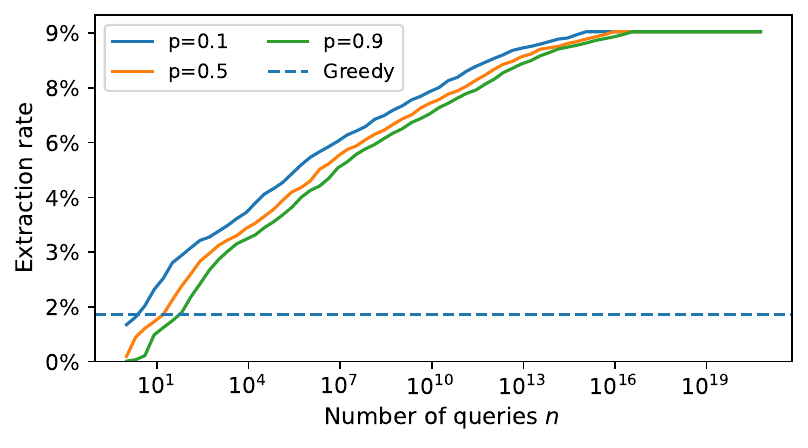}
  \end{center}
\vspace{-.4cm}
  \caption{Maximum extraction (Enron, Pythia 2.8B).\looseness=-1}
\label{fig:max-extraction}
\vspace{-.4cm}
\end{figure}

\begin{figure}
\centering
  \begin{center}
\includegraphics[width=.95\linewidth]{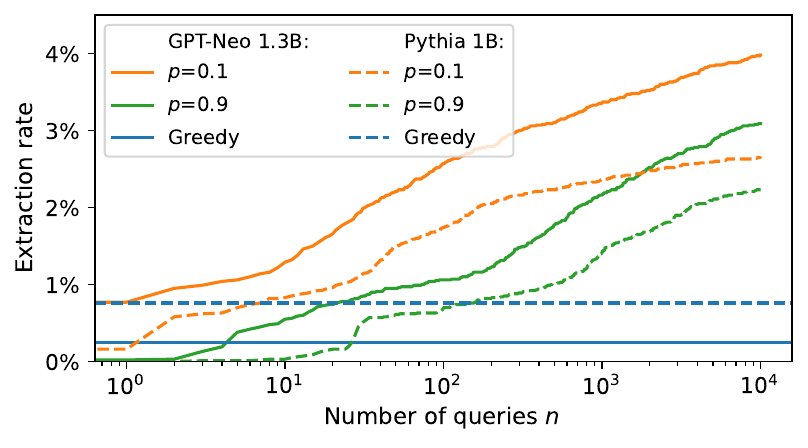}
  \end{center}
\vspace{-.4cm}
  \caption{Comparing extraction rates across two different models, GPT-Neo 1.3B and Pythia 1B, using Enron.}
\label{fig: gptneo_compare}
\vspace{-.4cm}
\end{figure}

In Figures~\ref{fig: model_sizes}-\ref{fig: gptneo_compare}, we plot extraction rates for different models,  according to various  choices of $n$ and $p$. 
The curves therefore represent the overall  $(n,p)$-discoverable extraction rate over the given set of training examples, drawn from the Enron dataset; 
they capture extraction risk as it varies in terms of $n$ and $p$ (Section~\ref{sec:def:def}). 
In contrast, the rate for  greedy-sampled discoverable extraction (\Cref{def:discoverable_extraction}) does not convey this type of information;   
the greedy rate is, of course, constant, because  greedy sampling is deterministic. 
In every experiment, there exist settings of $n$ and $p$ that catch cases of extraction that greedy sampling misses. 
This is clear because, on the same examples,  $(n,p)$-discoverable extraction rates surpass greedy extraction rates---even for relatively modest values of $n$ and $p$. 
Results across other model families, model sizes, datasets, and sampling schemes support these observations, and can be found in Appendices~\ref{sec:app_more_model_sizes}-\ref{app:sec:moreresults}. \looseness=-1 

We organize additional observations 
into two themes: 
reasoning about overall extraction risk for a single model (Figures~\ref{fig: model_sizes}~\&~\ref{fig:max-extraction}) and  comparisons of extraction risk between different models (Figure~\ref{fig: gptneo_compare}).

\custompar{Overall model extraction risk.} 
As expected from \Cref{eq:npmem},  the $(n,p)$-discoverable extraction rate appears to have a log-linear relationship with $n$ for all choices of $p$.   
As a result, there is a maximum amount of extraction that can be obtained for a particular model $f_\theta$ and sampling scheme $g_\phi$ over a given set of training data, regardless of the choice of $p$.  
That is, for each $p$, there exists an $n$ at which we obtain the worst-case  
extraction rate---an  
upper bound on the possible extraction risk.  
For top-$k$ sampling ($k\!=\!40$, $T\!=\!1$) with Pythia 2.8B  on Enron, this worst-case rate is 9.04\% (\Cref{fig:max-extraction}), which is nearly $7\times$ the greedy rate (1.3\%, \Cref{fig: topk2b}). 

Beyond worst-case extraction risk, important patterns emerge for different settings of $n$ and $p$. 
Recall that low-$p$ settings are the most permissive for deeming a target to be extractable: 
such settings are appropriate to consider when it may be a problem if there is even a small probability of generating the target at least once (e.g., for PII, see \Cref{sec:def:hps}). 
\Cref{fig: model_sizes} plots this setting for  $p\!=\!0.1$. 
Even at very small $n$, the $(n,p)$-extraction rate is higher than the greedy rate for both Pythia 2.8B (1.3\%) and 12B (3.07\%). 
That is, for low $p$, greedy sampling \emph{underestimates} extraction risk; 
this underestimate is large even for ($n\!>\!3$), and becomes significantly larger as $n$ increases.\looseness=-1 \looseness=-1 

From another perspective---the high-$p$, low-$n$ setting---greedy-sampled discoverable extraction can also be viewed to \emph{overestimate} extraction. 
Consider  
$p=0.999$, for which it is almost certain 
that a target sequence is generated at least once. 
While for large $n$, $p=0.999$ shows that the greedy rate significantly underestimates extraction, for small $n$ the $(n,p)$ rate is \emph{lower} than the greedy rate. 
For example, for Pythia 2.8B, the $(n,p)$ rate only approaches the greedy rate when $n\!>\!169$. 
Even for $p=0.5$, where it is effectively a coin flip that the target is generated at least once, the $(n,p)$ rate only approaches the greedy one when $n\!>\!17$.  
That is, if we only allow (e.g., rate limit) a relatively small number of queries, the $(n,p)$-discoverable extraction rate can be kept quite low---indeed, close to 0\% for high $p$ and $n\!<\!10$. 
From this perspective, greedy-sampled discoverable extraction---the metric computed in many model-release reports~\citep[e.g.,][]{reid2024gemini, grattafiori2024llama3herdmodels}---may give an overly pessimistic picture of the amount of extraction experienced by end users in practice.\looseness=-1 

\custompar{Extraction risk across models.} 
Measuring $(n, p)$-dis\-cov\-erable extraction also reveals trends  
across models that are not apparent for  greedy-sampled discoverable extraction.  
To show this, we compare extraction rates for Pythia 1B with GPT-Neo 1.3B~\citep{gptneo}, which is a similar-sized model also trained on the Pile (\Cref{fig: gptneo_compare}). 
We observe that Pythia 1B has a larger greedy  extraction rate than  
GPT-Neo 1.3B; 
however, for the same $p$ at \emph{at every} $n$, the $(n, p)$-discoverable extraction rate  
for GPT-Neo 1.3B is larger than for Pythia 1B. 
A practitioner who only measures  discoverable extraction with greedy sampling would  falsely conclude that Pythia 1B is at higher risk for extracting training data.  
In contrast, $(n,p)$-discoverable  extraction implies the opposite. 
By providing a more reliable estimate of extraction risk in terms of $n$ and $p$, our probabilistic measure  
facilitates better comparisons of extraction risk across models.\looseness=-1

\subsection{Evaluating model size and data repetition}\label{ssec: gap}

\begin{figure}[t]
\centering
    \includegraphics[width=.95\linewidth]{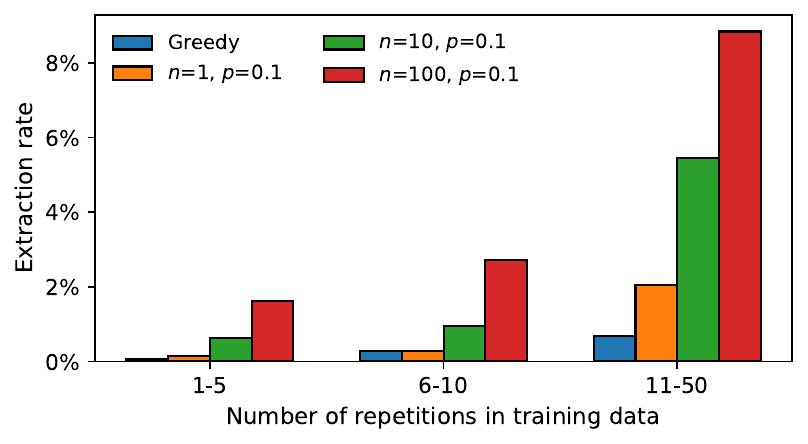}
  \vspace{-.2cm}
  \caption{Rates for greedy discoverable and $(n, p)$-discoverable extraction using top-$k$ sampling ($k=40$, $T=1$) 
  on Pythia 2.8B. 
  The $(n,p)$ rates exceed the greedy rate, and the gap widens with more repetitions.\looseness=-1}
\label{fig: topk2b_rep_data}
\vspace{-.4cm}
\end{figure}

It is well-known in the memorization literature that discoverable extraction rates tend to be higher for larger models~\citep{carlini2022quantifying, biderman2024emergent, lu2024scaling, tirumala2022memorization, mireshghallah2022empirical} and that repeated training-data examples are more likely to be extracted~\citep{lee2021deduplicating}. 
Our results confirm these trends:  
the greedy rates for Pythia 1B (\Cref{fig: gptneo_compare}), 2.8B (\Cref{fig: topk2b}), and 12B (\Cref{fig: topk12b}) are 0.76\%, 1.3\%, and 3.07\%, respectively; 
the greedy rate also increases as a function of repetitions (Figure~\ref{fig: topk2b_rep_data}). 
Our experiments with $(n,p)$-discoverable extraction follow the same patterns and also offer further insights.   
To the extent that greedy-based extraction attacks may be underestimating extraction at high $n$ and relatively low $p$, \emph{these underestimates are worse for larger models and for target data that has a higher number of repetitions}.\looseness=-1

\custompar{Model size.}
Similar to greedy-sampled discoverable extraction, $(n, p)$-discoverable extraction rates increase with model size for models in the same family.  
For example, at all $n$ and $p$, the $(n,p)$-discoverable extraction rates are higher for Pythia 12B compared to Pythia 2.8B (Figure~\ref{fig: model_sizes}). 
Further, the gap between the greedy-sampled discoverable extraction rate and  
$(n, p)$-discoverable extraction rates increases for larger models.
On the Enron training-data subset, 
for Pythia 2.8B, the gap between the greedy rate (1.3\%, see \Cref{fig: topk2b}) and maximum extraction rate  (9.04\%, see Figure~\ref{fig:max-extraction}) is 7.74\%. 
By comparison, the gap for Pythia 12B between the greedy (3.07\%, see Figure~\ref{fig: topk12b}) and maximum rate (16.07\%, see  Appendix~\ref{sec:app_more_model_sizes}) is 13\%. 
In a similar vein, 
it takes fewer queries for larger models to match the greedy rate. 
For example, at $p\!=\!0.9$, Pythia 1B
needs to generate $n\!=\!150$ sequences to exceed the greedy rate, while Pythia 12B  only needs to generate $n\!=\!40$ sequences.

\custompar{Example repetitions.}
The $(n,p)$-discoverable extraction rate also  increases as a function of training-example repetitions, and the gap between the greedy  and the $(n,p)$ rates is wider for more repetitions. 
For this experiment, depicted in Figure~\ref{fig: topk2b_rep_data}, we find phone numbers that are replicated within the Pile (Pythia's training dataset).   
We use each phone number as the target suffix, and the preceding text as the prefix to serve as the prompt.

\subsection{Validating $(n,p)$-discoverable extraction}\label{ssec:relative_risk}

A natural concern for  
using large $n$ and small $p$ is that this setting is too permissive to provide meaningful measurements of extraction.
That is, for sufficiently large $n$, a model might just happen to output the target suffix with low probability---even if that suffix was not memorized. 
If this is the case, then our measurements for $(n,p)$-discoverable extraction might mix together true instances of extracting memorized training data with instances of generating training data by happenstance; 
our measurements might not be valid estimates for memorization.\looseness=-1  

\begin{figure}[t]
\centering
    \includegraphics[width=.95\linewidth]{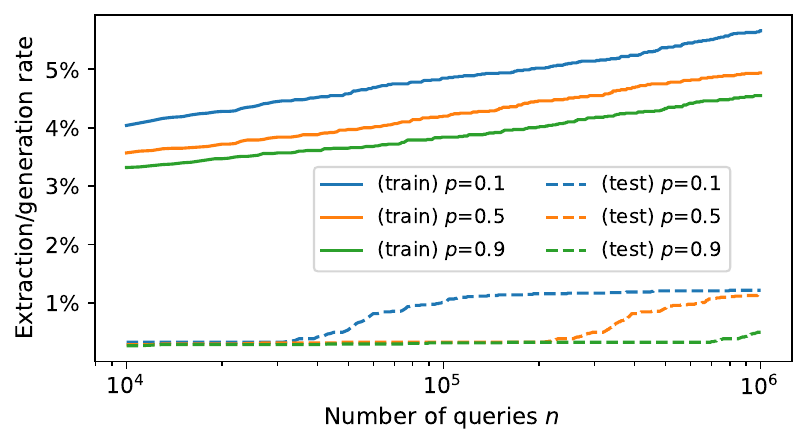}
\vspace{-.3cm}
  \caption{Validating training-data extraction. 
  For Pythia 2.8B, we compare extraction rates of training data (\textbf{train}) from Enron to the rates of generating verbatim unseen test data (\textbf{test}) from TREC 2007 Spam. 
  }
\label{fig: topk2btraintest}
\vspace{-.3cm}
\end{figure}

We investigate this possibility in Figure~\ref{fig: topk2btraintest}, and our results support that our measurements for $(n,p)$-discoverable extraction are indeed valid. 
For Pythia 2.8B, we compare extraction of training data from Enron to the 
generation of unseen test data. 
For the test data, we use  
$10,000$ emails from the TREC 2007 Spam classification dataset~\citep{bratko2006spam}. 
We find that, at all settings for $p$, the rate of generating test data is very low, and is significantly lower than the corresponding rate of generating training data. 
Even for $p=0.1$ and very large $n$, the test-data generation rate is less than 1\%---compared to over 5\% for training-data extraction. 
In fact, the test-data generation rate is effectively $0\%$ for $p\!=\!0.1$ until $n\!>\!50,000$.  
For $p\!=\!0.9$, it is 0.3\% even after $500,000$ queries, compared to 
4.4\% for the $(n,p)$-discoverable extraction rate---a difference of over an order of magnitude. 

In other words, for all settings of $p$, the number of queries $n$ needed to generate unseen test data is orders of magnitude larger than for generating training data.  
We find that it is generally  challenging to generate test data---even for low $p$---and especially in comparison to our measurements for extracting training data. 
This supports that, in our measurements of $(n,p)$-discoverable extraction, matches between training-example targets and generated suffixes are almost surely due to memorization.\looseness=-1

\section{Conclusion}\label{sec:discuss}

In this paper, we take a probabilistic perspective on measuring training-data extraction in language models.
This represents a significant departure from prior work and model-release reports, which tend to measure discoverable extraction: 
one-shot, yes-or-no determinations of extraction using deterministic greedy sampling. 
Instead, our measure for $(n,p)$-discoverable extraction captures a continuous notion of the risk of extracting a target example: 
we recast discoverable extraction in terms of the number of queries $n$ one would need to make to extract a target at least once with probability $p$ under a chosen sampling scheme. 
To conclude, we revisit three benefits of our probabilistic approach. 

\custompar{Reliable quantification of extraction.} 
Through extensive experimentation, we show that $(n,p)$-discoverable extraction provides more reliable measurement of extraction rates and facilitates more valid comparisons of extraction rates across models. 
Greedy-sampled discoverable extraction---the common approach in prior work---often significantly underestimates the overall rate of possible extraction (Sections~\ref{sec:def:def} \&~\ref{ssec: n_and_p}), with the degree of underestimation increasing for larger models (Section~\ref{ssec: gap}).   
However, for low query budgets $n$ (and even for relatively low $p$),  the $(n,p)$-discoverable extraction rate is often lower than the greedy rate. 
This suggests that 
prior reported greedy rates may also overestimate the amount of extraction experienced by end users in practice (Section~\ref{ssec: n_and_p}).\looseness=-1 

\custompar{No computational overhead.}
Importantly, we obtain these results with no additional cost compared to traditional discoverable extraction.
This is because, in practice,  
$(n,p)$-discoverable extraction can be computed with just one query  (Section~\ref{sec:def:hps}).
So, with no overhead, our measure yields more nuanced information about extraction risk.\looseness=-1

\custompar{Risks beyond overall extraction.} 
The results we present concern overall extraction rates; 
however, $(n,p)$-discoverable extraction can also be leveraged for finer-grained analysis of risks associated with different levels of data sensitivity (Section~\ref{sec:def:hps}). 
For example, while extracting generic phrases may not pose a significant risk, even rare PII leakages can be problematic. 
Our measure allows practitioners to adjust $n$ and $p$ based on the extraction-risk tolerance appropriate to specific contexts. 
Further, 
future research could adapt 
our work beyond memorization---to measure the risk of a model outputting any target sequence, such as those that reflect harmful or otherwise undesirable content.\looseness=-1

\section{Limitations}\label{sec: limitations}

Extraction is fundamentally challenging to measure; 
improvements come as researchers and practitioners discover new attacks~\citep[e.g.,][]{nasr2023scalable,nasr2025scalable,su2024extractingmemorizedtrainingdata}. 
We study a setting with a relatively benign adversary---one with API access only and limited side information---and defer study of more powerful adversaries to other work.
Additionally, we make no distinction between extracting different types of target sequences. 
In practice, some target sequences (e.g., PII) are significantly more sensitive than others (Sections~\ref{sec:def:hps}~\&~\ref{sec:discuss}), so their extraction rates should be measured separately. 
We only cursorily consider extraction of PII (specifically, phone numbers) for the purpose of measuring the effect of repetitions (Section~\ref{ssec:relative_risk}). 
We defer to future work to investigate extraction rates of different types of PII and other sensitive targets.\looseness=-1

\bibliography{template_refs}
\newpage
\onecolumn

\appendix
\clearpage
\section{More examples of how greedy sampling can miss extraction}\label{app: more_why}

\begin{figure*}[htbp!]
  \centering
\begin{subfigure}[t]{0.48\textwidth}
\centering
    \includegraphics[width=\linewidth]{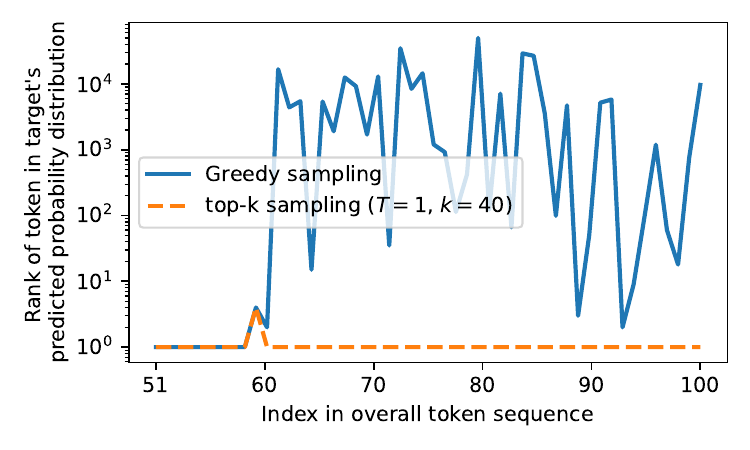}
      \vspace{-.6cm}
      \caption*{\textbf{Prefix:} 
      {\footnotesize
      \textit{``\textcolor{match}{Wygoda Sierakowska\escape{n}\escape{n}Wygoda Sierakowska  () is a village in the administrative district of Gmina Sierakowice, within Kartuzy County, Pomeranian Voivodeship}''}\\
      \textbf{Greedy suffix:} \textit{``\textcolor{match}{, in northern Poland. It lies approximately}  \textcolor{mismatch}{south-west of Sierakowice,  west of Kartuzy, and  west of the regional capital Gdańsk.\escape{n}\escape{n}For details of the history of the region,}''}\\ \textbf{Target suffix:} \textit{``\textcolor{match}{, in northern Poland. It lies approximately  east of Sierakowice,  west of Kartuzy, and  west of the regional capital Gdańsk.\escape{n}\escape{n}For details of the history of the region, see History}''}\\
      Edit distance between greedy and target suffix is $4\%$ (tokens) / $0.85\%$ (characters).
      Top-$k$ sampling outputs target suffix with probability $9.74\%$.}
      }
    \end{subfigure}\hfill
\begin{subfigure}[t]{0.48\textwidth}
\centering
    \includegraphics[width=\linewidth]{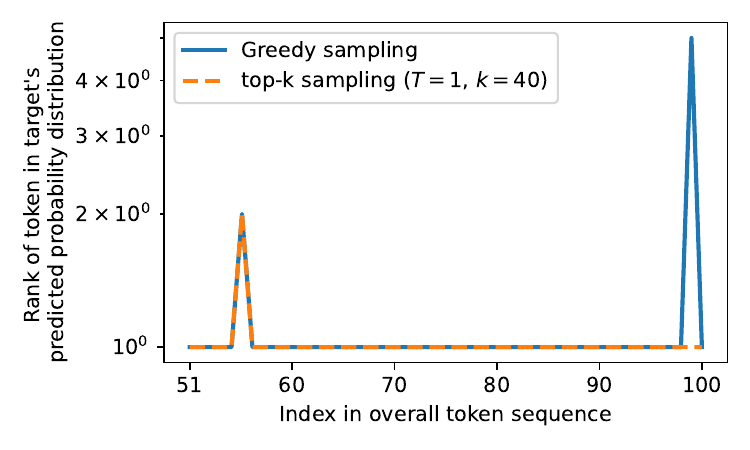}
        \captionsetup{width=\textwidth} 
        \vspace{-.6cm}
        \caption*{
        {\footnotesize
        \textbf{Prefix: }\textit{``\textcolor{match}{Yoshihiro Nikawadori\escape{n}\escape{n}Yoshihiro Nikawadori (\begin{CJK}{UTF8}{min}荷川取義浩\end{CJK}, Nikawadori Yoshihiro, born 4 December 1961) is a Japanese former handball}''}\\
        \textbf{Greedy suffix:} \textit{`` \textcolor{match}{player who competed in the} \textcolor{mismatch}{1984} \textcolor{match}{ Summer Olympics.\escape{n}\escape{n}References\escape{n}\escape{n}Category:1961 births\escape{n}Category:Living people\escape{n}Category:Japanese male handball players\escape{n}Category:Olympic handball players of Japan\escape{n}Category:Handball players at the} \textcolor{mismatch}{1984} \textcolor{match}{Summer}''}\\ 
        \textbf{Target suffix: } \textit{`` \textcolor{match}{player who competed in the 1988 Summer Olympics.\escape{n}\escape{n}References\escape{n}\escape{n}Category:1961 births\escape{n}Category:Living people\escape{n}Category:Japanese male handball players\escape{n}Category:Olympic handball players of Japan\escape{n}Category:Handball players at the 1988 Summer}''}\\
        Edit distance between greedy and target suffix is $10\%$ (tokens) /  $11.11\%$ (characters).
      Top-$k$ sampling outputs target suffix with probability $9.09\%$.\looseness=-1}}
\end{subfigure}
\begin{subfigure}[t]{0.48\textwidth}
\vspace{.3cm}
\centering
    \includegraphics[width=\linewidth]{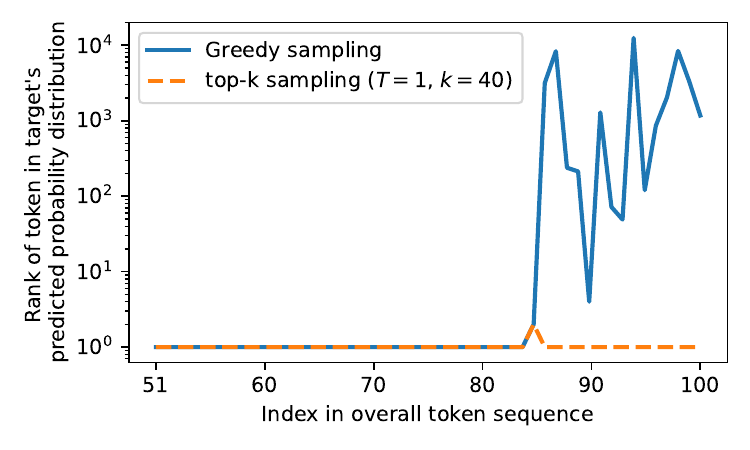}
        \captionsetup{width=\textwidth} 
        \vspace{-.6cm}
        \caption*{
        {\footnotesize
        \textbf{Prefix: }\textit{``\textcolor{match}{Greece at the 1984 Summer Paralympics\escape{n}\escape{n}Greece competed at the 1984 Summer Paralympics in Stoke Mandeville, Great Britain and New York City, United States. 3 competitors from Greece won no medals and so}''}\\
        \textbf{Greedy suffix:} 
        \textit{`` \textcolor{match}{did not place in the medal table.\escape{n}\escape{n}See also \escape{n} Greece at the Paralympics\escape{n} Greece at the 1984 Summer Olympics\escape{n}\escape{n}References \escape{n}\escape{n}Category:}\textcolor{mismatch}{Nations at the 1984 Summer Paralympics\escape{n}1984\escape{n}Summer Paral}''}\\
        \textbf{Target suffix: } 
        \textit{`` \textcolor{match}{did not place in the medal table.\escape{n}\escape{n}See also \escape{n} Greece at the Paralympics\escape{n} Greece at the 1984 Summer Olympics\escape{n}\escape{n}References \escape{n}\escape{n}Category:Greece at the Paralympics\escape{n}Category:1984 in Greek sport}''}\\
        Edit distance between greedy and target suffix is $22\%$ (tokens) / $21.81\%$ (characters).
      Top-$k$ sampling outputs target suffix with probability $14.93\%$.
        }}
        
\end{subfigure}\hfill
\begin{subfigure}[t]{0.48\textwidth}
\vspace{.3cm}
\centering
    \includegraphics[width=\linewidth]{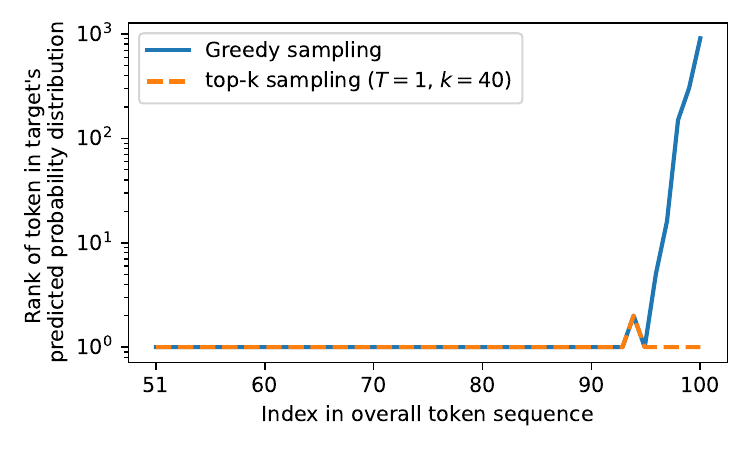}
        \vspace{-.6cm}
        \caption*{
        {\footnotesize
        \textbf{Prefix: }\textit{``\textcolor{match}{Conus patae\escape{n}\escape{n}Conus patae, common name Pat's cone, is a species of sea snail, a marine gastropod mollusk in the family Conidae, the cone snails and their allies.\escape{n}\escape{n}"}''}\\
        \textbf{Greedy suffix:} 
        \textit{``\textcolor{match}{Like all species within the genus Conus, these snails are predatory and venomous. They are capable of "stinging" humans, therefore live ones should be handled carefully or not at all.\escape{n}\escape{n}Description\escape{n}}\textcolor{mismatch}{The size of an adult}''}\\
        \textbf{Target suffix: } 
        \textit{``\textcolor{match}{Like all species within the genus Conus, these snails are predatory and venomous. They are capable of "stinging" humans, therefore live ones should be handled carefully or not at all.\escape{n}\escape{n}Distribution\escape{n}This species occurs in the}''}\\
        Edit distance between greedy and target suffix is $12\%$ (tokens) / $9.82\%$ (characters).
      Top-$k$ sampling outputs target suffix with probability $7.5\%$.
        }}
\end{subfigure}
\caption{Examples from the Pile (Wikipedia subset) of failures of greedy-sampled discoverable extraction from Pythia 12B. 
We prompt with the first 50 tokens in the target and test extraction with the subsequent 50 tokens. 
We also plot top-$k$ sampling  ($k\!=\!40$, $T\!=\!1$) generating the target suffix. 
We highlight matches with the target tokens in \textcolor{match}{blue} and mismatches in \textcolor{mismatch}{red}.
Determination of a (mis)match involves comparing tokens at the same index. 
Characters in the greedy suffix may match characters in the target suffix, but the indices can differ, causing a mismatch.\looseness=-1}
\label{fig:examples}
\end{figure*}

We give four more of examples in \Cref{fig:examples} of cases where greedy sampling misses extraction (Section~\ref{sec:def:def}). 

\FloatBarrier
\clearpage
\section{Comparison to other extraction, reconstruction, and memorization definitions}\label{app:sec:othermem}

In formulating our definition of probabilistic extraction (Section~\ref{sec:def}), our aim was to define something that 1) can easily be operationalized for production LLMs without needing to retrain multiple models, 2) roughly corresponds to the capabilities of a typical user, and 3) aligns with the risks posed by extraction. 
Numerous definitions of extraction, reconstruction, and memorization have been proposed, each with their own trade-offs that are not necessarily aligned with ours. 
We explore some of these definitions in this appendix.\looseness=-1 
\begin{figure*}[t]
\centering
\includegraphics[width=\linewidth]{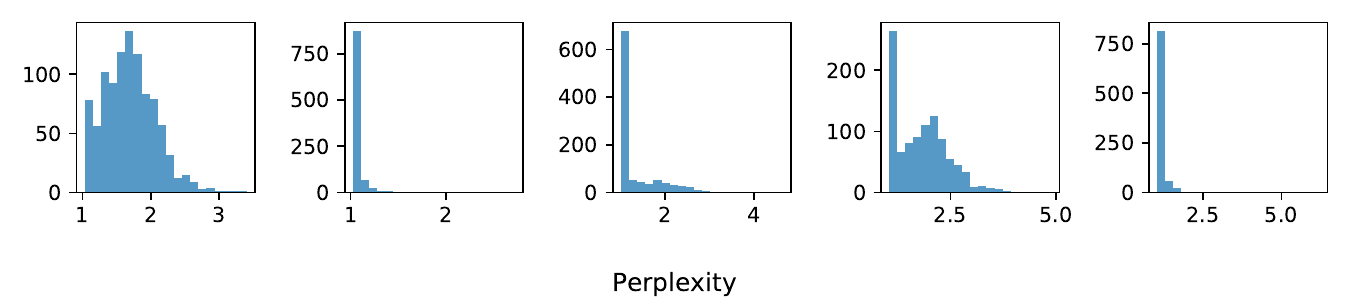}
        \caption{Distributions of perplexity values for $1000$ generated sequences, each  prompted using one of five  training-example prefixes. 
        The assumption in work by  \citet{carlini2019secret} that the empirical distributions are approximately (skewed) Gaussian, in order to estimate rank perplexity without sampling many times, is not appropriate for this setting. 
        Unlike in their setting for studying canary memorization, here, training examples are not  restricted to a bounded domain (e.g., phone numbers or social security numbers).\looseness=-1}
        \label{fig: secretshare}
\end{figure*}

\paragraph{Canary memorization.} 
As discussed in Section~\ref{sec:def:connections}, 
\citet{carlini2019secret} measure the unintended memorization of random strings (called \textbf{canaries}) inserted into the training dataset  via their \textbf{rank perplexities}. 
That is, an inserted canary has rank perplexity $i$ if the model perplexity on that canary is the $i$-th highest among all possible canaries that could have been sampled and inserted. 
(The possible canaries are sorted from lowest to highest perplexity.) 
This rank naturally corresponds to the number of guesses an adversary would need to make before correctly guessing the true canary, if the adversary guessed canaries in order from most to least likely (i.e., lowest to highest perplexity). 
This is a fairly natural setup for bounded sets of potential canaries; however, in our setting, it would be intractable to enumerate all possible token sequences of a given length.

\citet{carlini2019secret} use a (skewed) Gaussian approximation to model the expected distribution of perplexity values over a number of sequences, such that one can estimate rank perplexity  of a target without having to sample a large number of sequences.
The assumption of a Gaussian approximation being a good distributional fit is tailored to format-constrained canaries.
Even though this seems to align in principle with our $n$-query setup, in \Cref{fig: secretshare}, we show that this approach is not a good fit for our setting. 
We consider five different training examples from the Enron dataset, which is contained in Pythia's training dataset~\citep{gao2020pile}. 
For each example, we use its prefix to produce $1000$ generations, and we plot the perplexity distribution. 
Clearly, a (skewed) Gaussian is a relatively poor fit in all cases. 
Importantly, the empirical distributions vary from example to example; 
one could not estimate a distribution for one training example and expect this distribution to be a good fit for another example.
This makes estimating rank perplexity  challenging in our general, non-format-constrained   setting.

\paragraph{Information-theoretic definitions.} 
A number of memorization definitions have an information-theoretic flavor that rigorously capture a model's dependence on its training data. Unfortunately, estimating these dependencies often requires training many models, which is infeasible in our LLM setting.

\citet{brown2021memorization} measure the mutual information between the training dataset and the model output, conditioned on the data-generating distribution. 
This definition neatly captures the essence of memorization: 
the amount of information that is contained about the dataset in the model,  which is not a property of the underlying data distribution. 
They present two experiments as a proof-of-concept of their lower bounds. 
They sample the training data according to the specific learning task that exhibits their information-theoretic lower bounds, and then attack a logistic regression model and a single-hidden-layer feed forward network trained on this data. 
Importantly, their bounds require knowing the data-generating distribution.\looseness=-1

Other definitions similarly consider a counterfactual approach  to memorization, which requires estimating the model's performance with and without a particular training example~\citep{feldman2020neural, zhang2023counterfactual}. 
While~\citet{zhang2023counterfactual} use subsampling to reduce the number of models that need to be trained, they still require on the order of hundreds of models to get an estimate of  counterfactual memorization of the training data.

\paragraph{Memorization as compression.} 
Some works conceive of memorization as compression~\citep{lee2023talkin,cooper2023report,cooper2024files, franceschelli2024trainingfoundationmodelsdata, schwarzschild2024rethinking}.
As a concrete example, \citet{schwarzschild2024rethinking} study memorization by measuring the compression properties of an LLM with respect to a training example. 
They measure the length of the smallest prompt that will generate the target example in question as an indication of whether the material was memorized. 
In our setting, though, we are not necessarily concerned with extraction being proportional to the length of the prompt. 
One could imagine a very short but unusual prefix that reliably extracts a target,  versus a typical but longer prefix doing the same. 
Clearly, when measuring the risk of extraction by a regular user, the latter prompt is at higher risk of revealing the target even though it is longer, so this definition does not align with our setting.

\paragraph{Prompt engineering to measure extraction.} 

There is another line of work that uses clever prompting strategies to extract information from a target LLM. 
For example, \citet{kassem2024alpaca} use one LLM to find prompts that elicit extraction in another.
\citet{wang2024unlocking} construct prefix-dependent soft prompts to extract a given suffix. Both attacks involve optimizing over multiple prompts for each target sequence, which is more computationally expensive than our approach.

\paragraph{Probabilistic extraction.} 
Several other works study ideas related to probabilistic extraction. 
\citet{tiwari2025sequencelevelleakagerisktraining} published work shortly after our own, which is most directly related to what we study in this paper. 
They develop a similar motivation to our own, recognizing that measuring extraction beyond a one-shot approach with greedy sampling is 1) more reflective of realistic settings and 2) likely to show that prior work on discoverable extraction that takes a one-shot approach underestimates the true risk of extraction. 
Similar to our work, they examine several sampling  strategies and models (Llama and OPT). 
They also discuss a similar strategy for quantifying non-verbatim probabilistic memorization, where the extracted sequence is within a given edit distance of the target. They categorize their findings into six different patterns.
In contrast, we show that it is possible to derive a relationship between $n$ and $p$ for $(n,p)$-discoverable extraction in just one query  (Section~\ref{sec:def:hps}). 
We additionally run experiments to confirm that our extraction rates are capturing valid instances of memorization, as opposed to (potentially low-probability) generation of any target suffixes for large query budgets  (Section~\ref{ssec:relative_risk}).\looseness=-1

\citet{stock2022defendingreconstructionattacksrenyi} study reconstruction attacks, and show that 
R\'enyi differential privacy with DP-SGD can protect against reconstructing (i.e., extracting) canaries (which they also refer to as secrets) in GPT-2 fine-tuning experiments. 
They develop a ``lazy sampling'' approach to identify target sequences that are likely to be reconstructed. 
While we use probability $p$ to reason about the confidence that a target is in the set of $n$ generations, their analysis uses expectations. 

\citet{kim2023propileprobingprivacyleakage} develop black-box and white-box probing methods for studying PII-leakage risk in LLMs. 
They assess PII leakage with several metrics, including quantifying the likelihood of a particular $r$-length, PII-containing target being generated in response to a $q$-length prompt. 
With their particular focus on PII and data subjects, they also introduce a related metric, which quantifies the percentage of data subjects whose PII is leaked within $k$ queries to the model (with probability of leakage being greater than $\frac{1}{k}$).\looseness=-1

\citet{nakka2024piicompassguidingllmtraining} also specifically study PII leakage from LLMs.
In a subset of their experiments, they evaluate manual template-based prompts using top-$k$ sampling and $128$ repeated queries. 
For two of the manual templates, they observe an increase in extraction rate (compared to a single query). 
While this type of analysis is of a similar flavor to ours, it is not the focus of their work and is limited in scope.\looseness=-1 

\citet{cooper2024files} suggest measuring the probability of extraction for particular sequences, rather than relying on one-shot extraction attempts. 
However, building on their prior work~\citep{cooper2022lawless}, their focus is on why such an approach would be interesting for a legal audience.
They do not investigate this idea with tools in machine learning, as we do here. 

\section{Experimental setup}\label{app:sec:setup}

We provide additional details on the models we analyze, the datasets from which we draw prefixes and suffixes to test, and the procedure for using Equation (\ref{eq:npmem}) to compute the relationship between $n$ and $p$ for a given training example $\vz$ using just one query. 

\subsection{Models}

We use GPT-Neo 1.3B~\citep{gptneo}, the Pythia model  family~\citep{biderman2023pythia} (1B, 2.8B, 6.9B, and 12B), the Llama 1 model family (7B, 13B) ~\citep{touvron2023llamaopenefficientfoundation}, and the OPT model family (350M, 1.3B, 2.7B, 6.7B) for our experiments. 
Both GPT-Neo 1.3B and the Pythia model family are open-weight and open-data models trained by EleutherAI on the Pile~\citep{gao2020pile}. 
The Llama and OPT families are open-weight models released by Meta; 
while the Llama paper includes information on the training-data mix, full knowledge of the Llama's and OPT's training datasets is not publicly available.\looseness=-1

In the main paper and Appendices~\ref{sec:app_more_model_sizes}-\ref{app:sec:perplexity}, we primarily include experiments with the Pythia model family on the Enron training-data subset. 
We also run an experiment with GPT-Neo 1.3B on Enron, in~\Cref{fig: gptneo_compare} and run our Equation (\ref{eq:npmem}) verification experiment with Pythia 6.9B on Wikipedia data in Figure~\ref{fig:emp_p}. 
In Appendix~\ref{app:sec:moreresults}, we include experiments for all model families on other training-data subsets and proxies.\looseness=-1 

\subsection{Datasets}

We use a variety of datasets, from which we draw examples that we divide into prefixes for prompts and suffixes to check against generations. 
We briefly discuss training datasets and proxy datasets according to their corresponding  models.\looseness=-1 

\paragraph{Pythia and GPT-Neo 1.3B.} These open-weight, open-data models were trained on The Pile~\citep{gao2020pile}. 
For these models, we test for extraction on different training-data subsets that are included in the Pile: \textbf{Enron}, \textbf{Wikipedia}, and \textbf{GitHub} subsets. 
For more information on The Pile, we refer the reader to \href{ https://github.com/EleutherAI/the-pile?tab=readme-ov-file}{EleutherAI}. 
For the \href{https://github.com/noanabeshima/wikipedia-downloader}{Wikipedia} and \href{https://github.com/EleutherAI/github-downloader}{GitHub} subsets, we relied on the linked downloaders. 
As noted above, in the main paper we primarily run experiments on the Enron subset. 
In Appendix~\ref{app:sec:moreresults}, we run experiments on Pythia models for Wikipedia and GitHub. 
All three datasets have $10,000$ examples.

\paragraph{Llama and OPT.}
We do not know the exact training datasets for Llama and OPT, so we uses proxies for drawing examples to test for extraction. 
Since we know that Llama relied on Common Crawl data for training, we use $10,000$ examples drawn from Common Crawl. 
It is, of course, possible that the examples we use were  not contained in the OPT or Llama training datasets. 
We expect that our estimates for extraction to be lower for these models. 

\paragraph{Test dataset.} 
For our experiment for validating $(n,p)$-discoverable extraction by comparing to generation of test data in Pythia 2.8B, we use the Trek 2007 Spam classification dataset~\citep{bratko2006spam}.\looseness=-1 

\subsection{Computing $(n,p)$-discoverable extraction}\label{app:sec:setup:computing}

In the main paper, we compute $(n,p)$-discoverable extraction using two different procedures. 
For the most part, we use Equation (\ref{eq:npmem}), which provides the relationship between $n$ and $p$ for an example $\vz$ using only one query to the model to find $p_\vz$. 
This provides an efficient approximation of actually sampling $n$ queries to compute extraction probabilities. 
We discuss both procedures in a bit more detail here. 
We use the method involving just one query for all experiments except for Figure~\ref{fig:emp_p}, where we verify that this procedure matches empirically with using $n$ queries.\looseness=-1 

\paragraph{Computations of probabilistic extraction with just one query.}

To use Equation (\ref{eq:npmem}) to compute the relationship between $n$ and $p$, we use one query to compute $p_\vz$ for a given training example $\vz$. 
To do so, we sequentially feed the example $\vz$ into the model $f_\theta$ one token at a time. 
For each token, the model produces a conditional probability distribution over the vocabulary of the next token given the previous tokens. 
These conditional probabilities represent the base likelihood of observing a particular token at each position in the sequence. 
We use the sampling scheme $g_\phi$ to post-process these conditional probabilities. 
For example, for top-$40$ sampling and $T=1$, we set all token probabilities, except for the 40 tokens with the highest probabilities, to 0, normalize the 40 remaining non-zero token probabilities, and use Equation (\ref{eq:temp}) with $T=1$ to re-weight the model $f_\theta$'s base conditional probabilities. 
We then select the next actual token in $\vz$; 
we add this token's conditional $\log$ probability to a running sum, and repeat the process for the next token, and so on. 
Altogether, we compute the overall $\log$ likelihood of the entire sequence $\vz$ under both the model $f_\theta$ and the sampling scheme $g_\phi$ by adding the transformed conditional $\log$ probabilities for each token in the sequence. 
We use this overall $\log$ likelihood to produce  $p_\vz$, which we can then use with Equation (\ref{eq:npmem}) to derive $n$ for any fixed $p$, or vice versa.\looseness=-1 

\paragraph{Testing different prefix and suffix length with no additional queries.} 
Note that, for the above procedure, we can actually test \emph{any} prefix and suffix length for $\vz$ using just one query.
When we compute $p_\vz$, this depends on adding together the conditional $\log$ probability for each token. For a given suffix $\vz_{a+1:a+k}$, this just involves adding together the conditional $\log$ probabilities of the tokens at indices ${a+1, \ldots, a+k}$ after having already processed the prefix $\vz_{1:a}$ through the model. 
If we keep track of the per-index conditional $\log$ probabilities for the whole sequence, we can just change which ones we sum together (i.e., change $a$ and $k$) on the fly; 
with just one query to the model, we obtain enough information to compute $p_\vz$ for different prefix and suffix lengths, and could examine how these differences impact extraction.\looseness=-1

Aside from the other benefits of our approach with respect to surfacing extraction risk, this is another clear benefit in comparison to traditional discoverable extraction (Definition~\ref{def:discoverable_extraction}). 
Since discoverable extraction has traditionally involved yes-or-no determinations that just compare the generated output with the target suffix, testing for different prefix and suffix lengths entails making different queries to the model using each prefix as a prompt. 
Here, by examining probabilities, we are able to efficiently compute different measurements of extraction with greater flexibility, and with no additional queries to the model. 

\paragraph{$n$-shot computations of probabilistic extraction.} 
The above procedure works in one query by computing $p_\vz$ directly for the training example $\vz$. 
It does not actually sample different generations using a prefix $\vz^t_{1:a}$ and analyzing output  suffixes $\vz^o_{a+1:a+k}$ to see if they match the target suffix $\vz^t_{a+1:a+k}$.  
We can use $n$ queries to directly approximate $p_\vz$:
$n$ independent times, we prompt the model $f_\theta$ with $\vz_{1:a}$ and then sample with $g_\phi$; 
for this given $n$, we compute
\vspace{-.2cm}
\begin{align}
\label{eq:hatnp_mem}
\hat{p}_\vz=\frac{\sum_{w \in [n]}\1[\vz^{o,w}_{a+1:a+k} = \vz^t_{a+1:a+k}]}{n},
\end{align} 
\vspace{-.5cm}

\noindent which we then use to compute the corresponding $p$ similarly to Equation (\ref{eq:npmem}), using $\hat{p}_\vz$ in place of $p_\vz$, i.e.,\looseness=-1
\vspace{-.2cm}
\begin{align*}
    1 - (1 - \hat{p}_\vz)^n \geq p.
\end{align*}
\section{Extending to non-verbation extraction of targets}\label{app:sec:nonverbatim}

In Section~\ref{sec:def:apprx}, we extend the definition of $(n,p)$-discoverable extraction (Definition~\ref{def:np_discoverable_extraction}) from capturing only verbatim extraction of targets to also account for non-verbatim extraction of targets. 
We introduce a general variation of our definition (Definition~\ref{def:enp_discoverable_extraction}) that includes a distance function $\mathsf{{dist}}\!\!: \sV^k \times \sV^k \rightarrow \R_{\geq 0}$ and an $\epsilon \in \R_{>0}$ hyperparameter for the maximum allowable 
distance between the generated text and the target, such that the generated text would still count as a successful extraction. 

We deliberately leave this definition general, with the choice of distance function $\mathsf{{dist}}$ and $\epsilon$ up to the user. 
In this appendix, we discuss an instantiation of Definition~\ref{def:enp_discoverable_extraction} for the Hamming distance (Appendix~\ref{app:sec:nonverbatim:hamming}). 
In relation to this discussion, we then briefly show  how the definition could be modified for other edit-distance metrics  (Appendix~\ref{app:sec:nonverbatim:other}). 

\subsection{An example instantiation with the Hamming distance}\label{app:sec:nonverbatim:hamming} 

In our definition for $(\epsilon,n,p)$-discoverable extraction (Definition~\ref{def:enp_discoverable_extraction}), the implemented check for verifying extraction is modified from the check in the verbatim definition.
Rather than checking for exact equality between the generated text and the target, we instead check if the generated suffix is a member of the set of all suffixes that are within $\epsilon$ distance (where distance is computed with $\mathsf{{dist}}$) of the target. 
Here, we discuss a concrete instantiation of this definition with the Hamming distance---i.e., suffixes that are within $\epsilon$ token substitutions of the target suffix. 

That is, for $\epsilon \! \geq \! 1$, we define the set of $k$-length sequences    
$\sS_\epsilon(\vb) = \{\vc \mid \textsf{\footnotesize{HammingDistance}}(\vb, \vc) \leq \epsilon\}$; 
these are the possible suffixes.   
We check the generated text for membership in the set  $\sT_\epsilon = \{\vz_{1:a} \!\!\parallel\!\! \vs \mid \vs \in   \sS_\epsilon(\vz_{a+1:a+k}) \}$, where $\vz_{1:a}$ is the target prefix (and the prompt) and $\sS_\epsilon(\vz_{a+1:a+k})$ is the set of suffixes within $\epsilon$ distance of the verbatim target suffix $\vz_{a+1:a+k}$.  
In our context, the maximum number of token substitutions is $k$; this would be the maximum possible amount to set $\epsilon$. 
(Though, of course, $k$ would not be a meaningful choice for $\epsilon$, as this would result in any $k$-length sequence constituting a match.) 
We provide additional observations about this definition.\looseness=-1 

\paragraph{Cost of enumerating non-verbatim suffixes.}  
While in principle this is only a slight modification of our definition for probabilistic extraction, in practice it is significantly more computationally expensive to compute: 
for a given $\epsilon$, $k$-sized suffixes, and vocabulary $\sV$, there are $\binom{k}{\epsilon} \cdot \left| \sV \right|^\epsilon$ substitution-edit suffixes to enumerate and consider as potential candidates for valid extraction. 
Even for just $\epsilon = 1$, $50$-token suffixes and vocabulary $|\sV|=32,000$ (Llama's vocabulary size) yield $\binom{50}{1} \cdot 32,000^1 =1,600,000$ suffixes. 
For $\epsilon=2$, there are $\binom{50}{2} \cdot 32,000^2 = 39,200,000$ suffixes. 
(And, of course, for $\epsilon \leq 2$, we would need to enumerate and consider the union of these two sets.) 

\paragraph{No efficient, direct analogue for one-query computations.}  
Unlike for $(n,p)$-discoverable extraction, there is no direct analogue for Equation (\ref{eq:npmem}) that can compute $(\epsilon, n, p)$-discoverable extraction in only one query.
Recall that, in Equation (\ref{eq:npmem}), $p_\vz$ is the probability of generating a suffix $\vz_{a+1:a+k}$ for prefix $\vz_{1:a}$, given a  model, sampling scheme, and example $\vz=\vz_{1:a} \!\parallel\!\vz_{a+1:a+k}$.  
This means that  
the probability of \emph{not} generating $\vz_{a+1:a+k}$ in a single draw from the sampling scheme is $1-p_\vz$, and the probability of not generating $\vz_{a+1:a+k}$ in $n$ independent draws 
is $(1-p_\vz)^n$. 
Therefore, example $\vz$ is $(n,p)$-discoverably extractable for $n$ and $p$ that satisfy $1-(1-p_\vz)^n \geq p$.

For $(\epsilon, n, p)$-discoverable extraction, we could derive an analogous expression in terms of not generating \emph{any} of the suffixes in $\sS_\epsilon$. 
This would mean computing the probability of \emph{not} generating any $\vz_\epsilon \in \sS_\epsilon$ in a single draw from the sampling scheme is $1-\sum_{\vz_\epsilon \in \sS_\epsilon} p_{\vz_\epsilon}$, and the probability of not generating any  $\vz_\epsilon \in \sS_\epsilon$ in $n$ independent draws 
is $(1-\sum_{\vz_\epsilon \in \sS_\epsilon} p_{\vz_\epsilon})^n$. 
And so, example $\vz$ is $(\epsilon, n,p)$-discoverably extractable for the given $\epsilon$, $n$, and $p$ that satisfy
\begin{align}
\label{eq:enp_mem}
1-(1-\sum_{\vz_\epsilon \in \sS_\epsilon} p_{\vz_\epsilon})^n \geq p.
\end{align}
\noindent This computation would require us to enumerate every $\vz_\epsilon \in \sS_\epsilon$, and evaluate its associated probability $p_{\vz_\epsilon}$!\looseness=-1 

\paragraph{More efficient alternatives.}
Instead, we note that the empirical procedure for computing $(\epsilon, n, p)$-discoverable extraction can be made more efficient than computing this theoretical, closed-form expression. 
In our empirical procedure for $(n,p)$-discoverable extraction, for each example $\vz$, we generate $n$ sequences; 
we compute an empirical $\hat{p}_\vz$ with Equation (\ref{eq:hatnp_mem}), which is the fraction of $n$ outputs that match the target suffix (Section~\ref{sec:def:hps}, Appendix~\ref{app:sec:setup:computing}).
For this given $n$ and estimated $\hat{p}_\vz$, we can use a similar procedure to Equation (\ref{eq:npmem}) to compute $p$ for the given model $f_\theta$ and sampling scheme $g_\phi$. 

We can follow a similar logic here for an approximation of Equation (\ref{eq:enp_mem}). 
For each of the $n$ generated suffixes, we can compute the Hamming distance with the target suffix of $\vz$, and count the suffixes that are within $\epsilon$ distance as successfully extracted.
This serves as an approximation $\hat{p}_{\vz_\epsilon}$. 
That is, $n$ independent times, we prompt the model $f_\theta$ with $\vz_{1:a}$ and then sample with $g_\phi$; 
for this $n$ and chosen $\epsilon$, we compute
\begin{align}
\label{eq:approxpze}
\hat{p}_{\vz_\epsilon}=\frac{\sum_{w \in [n]}\1\big[\textsf{\small{HammingDistance}}(\vz^{o,w}_{a+1:a+k},\vz^t_{a+1:a+k})\leq \epsilon\big]}{n}.
\end{align}
We can use this combined estimate of $\hat{p}_{\vz_\epsilon}$ to compute the corresponding $p$ similarly to Equation (\ref{eq:npmem}), using $\hat{p}_{\vz_\epsilon}$ in place of $p_\vz$. Or, put differently, we would use  $\hat{p}_{\vz_\epsilon}$ as our approximation in place of  $\sum_{\vz_\epsilon \in \sS_\epsilon} p_{\vz_\epsilon}$ in Equation (\ref{eq:enp_mem}). 
For this approximation, we do not have to enumerate the whole set of suffixes $\sS_\epsilon$:  
we can lazily check if the suffixes we generate are members of $\sS_\epsilon$ one at a time, by computing the Hamming distance for each one with the target suffix and seeing if it is within $\epsilon$, i.e., 
\begin{align}
\label{eq:approxenp}
    1 - (1 - \hat{p}_{\vz_\epsilon})^n \geq p. 
\end{align}
Of course, this will only give us partial coverage of the (potentially very large) set $\sS_\epsilon$; 
for small $n$, our combined approximation of $\hat{p}_{\vz_\epsilon}$ may not be a high-quality stand-in for $\sum_{\vz_\epsilon \in \sS_\epsilon} p_{\vz_\epsilon}$ in Equation (\ref{eq:enp_mem}). For small $n$, we can instead estimate $\hat{p}_{\vz_\epsilon}$ by drawing $m > n$ examples and then compute $1 - (1 - \hat{p}_{\vz_\epsilon})^n$ as before.\looseness=-1

\subsection{Using other distances}\label{app:sec:nonverbatim:other} 

We note that other distances could be used instead of the Hamming distance in Equation (\ref{eq:approxpze}), in order to compute the estimate $\hat{\vp}_{\vz_\epsilon}$ that is used in Equation (\ref{eq:approxenp}) to derive the relationship for any $n$ and $p$. 
Indeed, any $\mathsf{{dist}}$ function that satisfies the definition $\sV^k \times \sV^k \rightarrow \R_{\geq 0}$ could be used. 
For example, we could use the Levenshtein distance; in addition to considering substitution edits, this distance also counts insertions and deletions. 
As a result, it handles token-index-shifted sequences differently than the Hamming distance for equivalent-length sequences (i.e., such sequences can have a lower Levenshtein distance than Hamming distance). 
We could also consider the normalized edit distance (often computed with the Levenshtein distance as the $\mathsf{EditDistance}$ metric), as in \citet{lee2021deduplicating}. 
For $\vb, \vc \in \sV^k$, 
\begin{align*}
    \mathsf{EditDistance}_{\textsf{norm}}(\vb, \vc) = \frac{\mathsf{EditDistance}(\vb, \vc)}{\max (|\vb|,|\vc|)}.
\end{align*}

\noindent \citet{lee2021deduplicating} also consider the Jaccard distance, which similarly could be used in Equation (\ref{eq:approxpze}). 
Similarity scores, like the BLEU score, could be turned into a distance metric and then also be used as the $\mathsf{{dist}}$ function. (For BLEU score, this would be $1 - \text{BLEU score}$.) 

\clearpage
\section{Experiments with more Pythia model sizes on Enron}\label{sec:app_more_model_sizes}

We expand upon the results presented in Section~\ref{ssec: n_and_p} for Pythia models, where we test extraction on the Enron training-data subset containing $10,000$ examples. 
Across most choices of $n$ and $p$, we find that when the number of model parameters doubles, extraction rates approximately double.

\begin{figure*}[htbp!]
\captionsetup[subfigure]{justification=centering}
  \centering
\begin{subfigure}[t]{0.49\textwidth}
\centering
    \includegraphics[width=1.\linewidth]{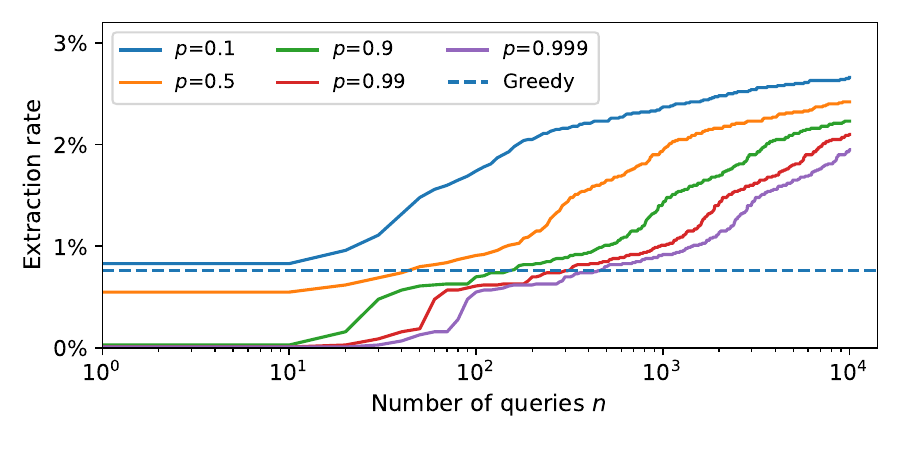}
        \caption{1B}
        \label{fig: app_topk1b}
\end{subfigure}
\hfill
\begin{subfigure}[t]{0.49\textwidth}
\centering
    \includegraphics[width=1.\linewidth]{figures/cooper-figures/plotting_data_top_k_40_enron_2b_seq_len_50_num_10000_unique.pdf}
        \caption{2.8B}
        \label{fig: app_topk2b_notmain}
\end{subfigure}
\begin{subfigure}[t]{0.49\textwidth}
\centering
    \includegraphics[width=1.\linewidth]{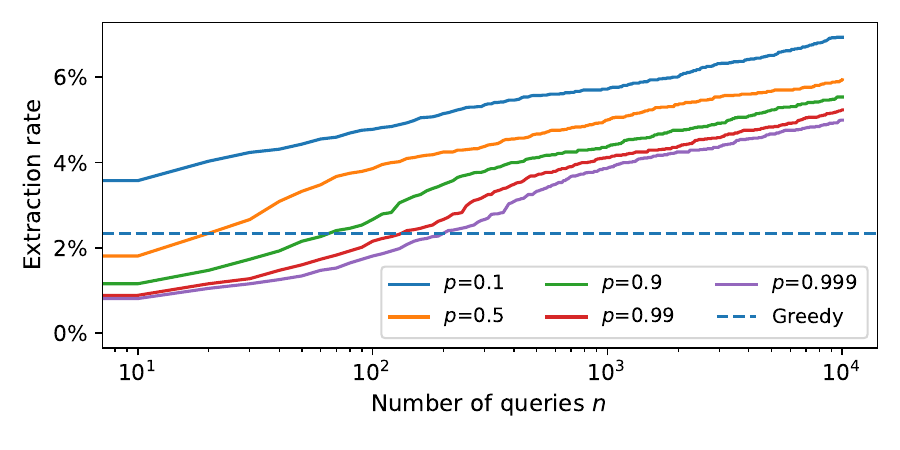}
        \caption{6.9B}
        \label{fig: app_topk6b}
\end{subfigure}
\hfill
\begin{subfigure}[t]{0.49\textwidth}
\centering
    \includegraphics[width=1.\linewidth]{figures/cooper-figures/plotting_data_top_k_40_enron_12b_seq_len_50_num_10000_unique.pdf}
        \caption{12B}
        \label{fig: app_topk12b}
\end{subfigure}
\caption{Illustrating $(n, p)$-discoverable extraction rates for different models in the Pythia family, using top-$k$ sampling ($k=40$, $T=1$) on the Enron dataset. 
This figure expands upon the results in Figure~\ref{fig: model_sizes}.}
\label{fig: app_model_sizes}
\end{figure*}

\begin{figure*}[htbp!]
\captionsetup[subfigure]{justification=centering}
  \centering
\begin{subfigure}[t]{0.49\textwidth}
\centering
    \includegraphics[width=1.\linewidth]{figures/cooper-figures/max-extract.pdf}
         \label{fig: app_max_2b}
        \caption{2.8B maxes out at 9.04\%.} 
\end{subfigure}
\hfill
\begin{subfigure}[t]{0.49\textwidth}
\centering
    \includegraphics[width=1.\linewidth]{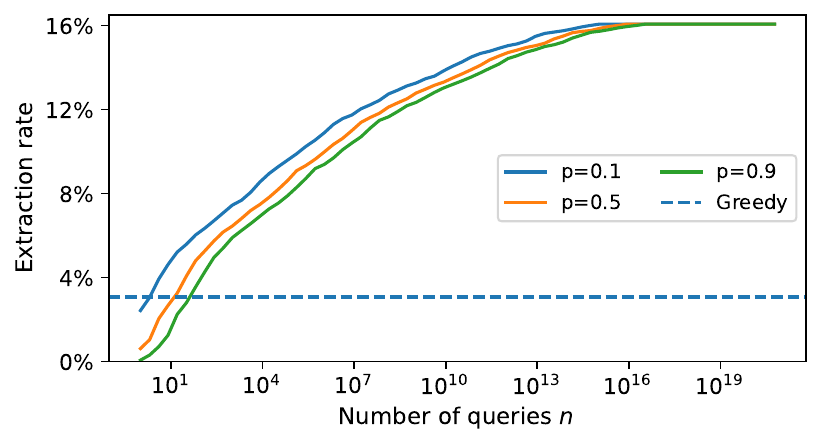}
        \label{fig: app_max12b}
        \caption{12B maxes out at 16.07\%.}
\end{subfigure}
\caption{Maximizing $(n, p)$-discoverable extraction rates for different models in the Pythia family, using top-$k$ sampling ($k=40$, $T=1$) on the Enron dataset. 
As discussed in Section~\ref{ssec: gap}, the greedy rate underestimates extraction more significantly for larger models. 
We can show this with the gap between the greedy and the maximum $(n,p)$-discoverable extraction rates. 
For Pythia 2.8B, the gap between the maximum rate (9.04\%) and the greedy rate (1.3\%, see \Cref{fig: app_topk2b_notmain}) is 7.74\%. 
For Pythia 12B, the gap between the maximum rate (16.07\%) and the greedy rate (3.07\%, see \Cref{fig: app_topk12b}) is 13\%.\looseness=-1}
\label{fig: app_max_extract}
\end{figure*}
\FloatBarrier
\clearpage
\section{Extraction rates under different sampling schemes for Pythia 2.8B on Enron}\label{sec:sampling_schemes}

\begin{figure*}[htbp!]
\captionsetup[subfigure]{justification=centering}
  \centering
\begin{subfigure}[t]{0.49\textwidth}
\centering
    \includegraphics[width=1.\linewidth]{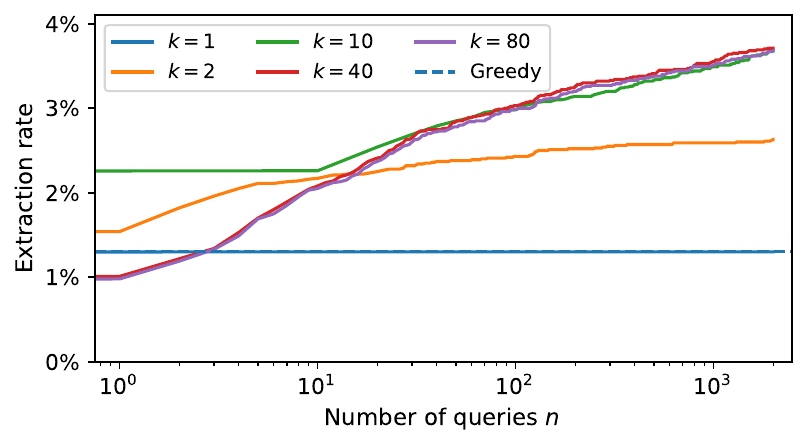}
        \caption{top-$k$, $p=0.1$}
        \label{fig: topkp01}
\end{subfigure}
\hfill
\begin{subfigure}[t]{0.49\textwidth}
\centering
    \includegraphics[width=1.\linewidth]{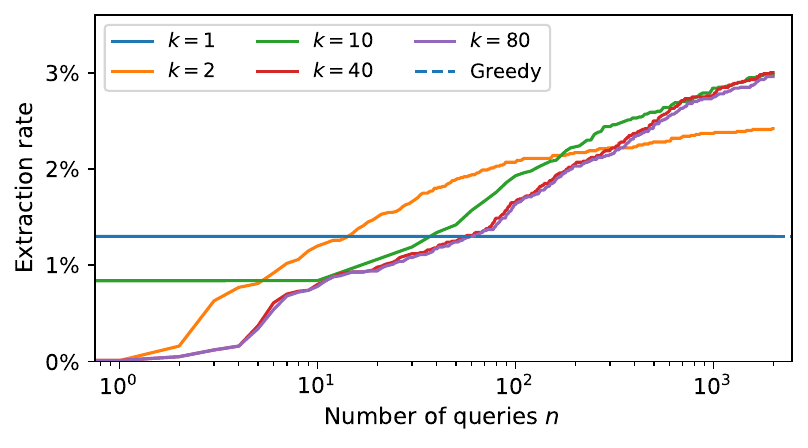}
        \caption{top-$k$, $p=0.9$}
        \label{fig: topkp09}
\end{subfigure}
\captionsetup[subfigure]{justification=centering}
  \centering
\begin{subfigure}[t]{0.49\textwidth}
\centering
    \includegraphics[width=1.\linewidth]{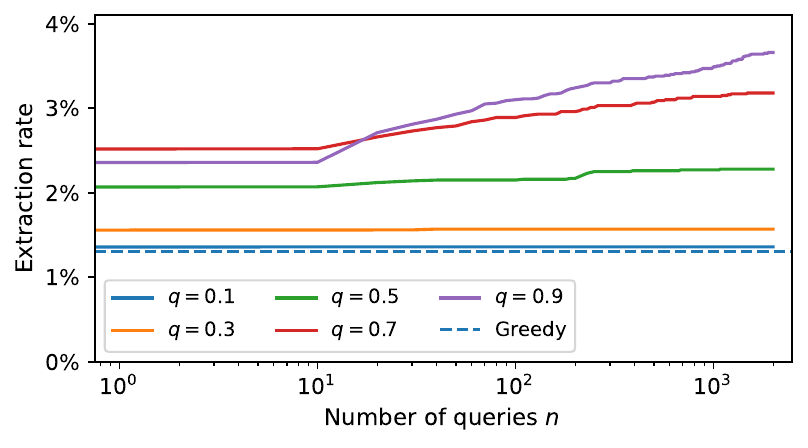}
        \caption{top-$q$, $p=0.1$}
        \label{fig: toppp01}
\end{subfigure}
\hfill
\begin{subfigure}[t]{0.49\textwidth}
\centering
    \includegraphics[width=1.\linewidth]{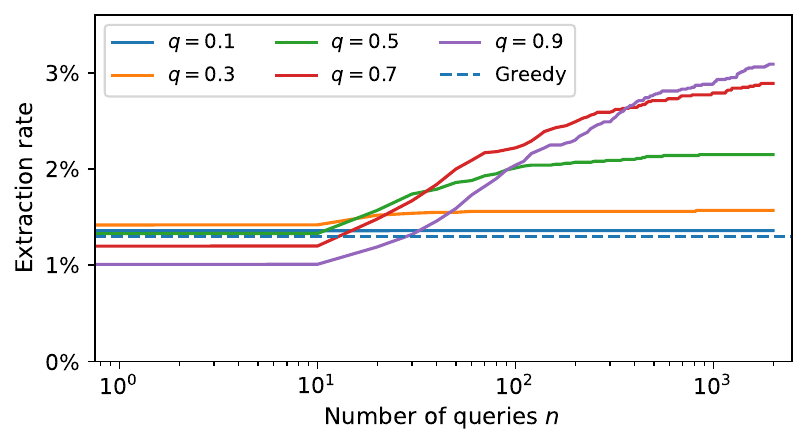}
        \caption{top-$q$, $p=0.9$}
        \label{fig: toppp09}
\end{subfigure}
\begin{subfigure}[t]{0.49\textwidth}
\centering
    \includegraphics[width=1.\linewidth]{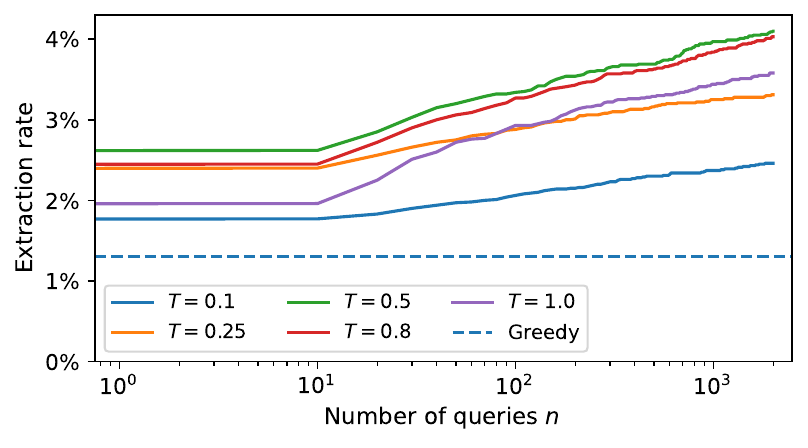}
        \caption{temperature ($T$), $p=0.1$}
        \label{fig: tempt01}
\end{subfigure}
\hfill
\begin{subfigure}[t]{0.49\textwidth}
\centering
    \includegraphics[width=1.\linewidth]{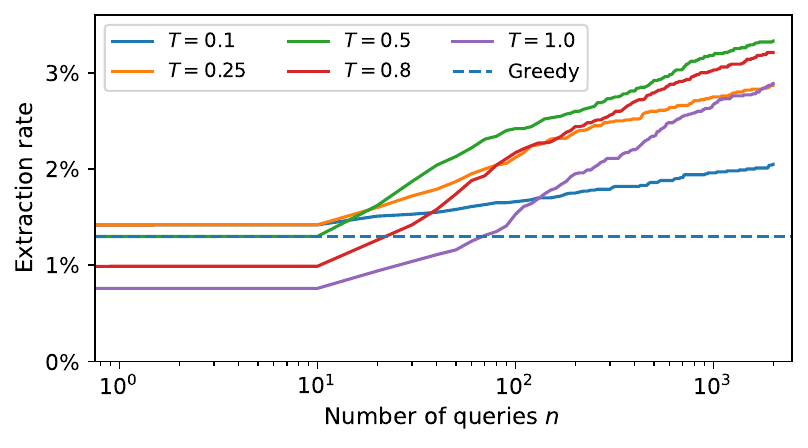}
        \caption{temperature ($T$), $p=0.9$}
        \label{fig: tempt09}
\end{subfigure}
\caption{Comparison of $(n, p)$-discoverable extraction rates for different sampling schemes. 
We fix $p$ to either $0.1$ or $0.9$. 
In Figures~\ref{fig: topkp01} and \ref{fig: topkp09} we vary $k$ in top-$k$ sampling, in Figures~\ref{fig: toppp01} and \ref{fig: toppp09} we vary $q$ in top-$q$ sampling, and in Figures~\ref{fig: tempt01} and \ref{fig: tempt09} we vary $T$ in temperature sampling. 
Generally, smaller $k$, $q$, and $T$ yield smaller extraction rates, but this also is dependent on both $n$ and $p$. 
However, for most hyperparameter values (even for small $n$), the extraction rate is above the greedy-sampled discoverable extraction rate.}
\label{fig: sampling_schemes}
\end{figure*}

Due to space constraints, in the main paper we focus on top-$k$ sampling as our sampling scheme $g_\phi$, with $k=40$ and $T=1$. 
Here, we provide more detailed results on extraction rates under different choices of $n$ and $p$ using different hyperparameters and sampling schemes for Pythia 2.8B on Enron ($10,000$ examples). 
We perform additional experiments for top-$k$ and sampling with different temperatures (\Cref{sec:prelim:sampling}).
We also include results for \textbf{nucleus sampling}, which we refer to as \textbf{top-$q$ sampling} in plots. 
Nucleus sampling is similar to top-$k$ sampling, but instead of keeping only the top-$k$ token probabilities, we retain (and normalize) the smallest subset of tokens such that their cumulative probability is at least $q\in(0, 1]$. 
Note that, in the literature, this is typically referred to as top-$p$ sampling; we relabel this as top-$q$ sampling to disambiguate with our use of $p$ in $(n,p)$-discoverable extraction.
The results are summarized in \Cref{fig: sampling_schemes}, where we vary the sampling-scheme-specific hyperparameters $k$, $q$, and $T$, respectively.\looseness=-1 

\paragraph{Top-$k$ sampling.} 
Increasing $k$ can substantially increase extraction rates even for moderate settings of $n$, and this effect becomes more pronounced with smaller values of $p$. 
In general, larger-$k$ settings often start out with lower extraction rates, but the rate of increase in the extraction rate (as a function of $n$) is more rapid, such that larger-$k$ extraction rates eventually exceed smaller-$k$ extraction rates. 
At small $n$, larger settings of $k$ may not surpass the greedy extraction rate (which, by definition, also matches $k=1$);
for larger $n$, it can be observed consistently that larger $k$ yields a higher extraction rate. 

For example, in \Cref{fig: topkp01}, where we fix $p=0.1$, the extraction rate at $k=2$ is approximately 1.5\% when $n=1$, which surpasses the $k=40$ and $k=80$ extraction rates (which are below the greedy rate of 1.3\%).
However, eventually these larger-$k$ rates surpass both the greedy and $k=2$ rate, and this difference only increases with $n$. 
A similar effect can be observed in \Cref{fig: topkp09}, where we fix $p=0.9$; 
however, in this case, all settings of $k>1$ start off below the greedy rate, and the differences in larger-$k$ rates are smaller.\looseness=-1 

In general, larger values of $k$ yield larger extraction rates, but this is affected by both $p$ and $n$. 
For small $p$, extraction rates for small $k$ are dominated by larger $k$;  for larger $p$, extraction rates for small $k$ dominate larger $k$, but this eventually reverses when $n$ becomes larger enough.

\paragraph{Top-$q$ sampling.}
Similar trends can be observed for top-$q$ sampling, in Figures~\ref{fig: toppp01} and \ref{fig: toppp09}, where larger values of $q$ result in larger extraction rates as $n$ increases. 
The gap between rates also increases as $n$ increases, though this occurs at a slower rate for larger $p$.

\paragraph{Random sampling.} 
Similar trends can also be observed for temperature sampling in Figures~\ref{fig: tempt01} and \ref{fig: tempt09}, though gaps between rates are more consistent at larger $n$. 

\paragraph{Why is there no strict ordering?} 
One may wonder why there is not a strict ordering of extraction rates for top-$k$, top-$q$, and temperature sampling, if rates are compared according scaling $k$, $q$, and $T$, respectively. 
As $k$, $q$, or $T$ is varied, token probabilities can either increase or decrease. 
For example in top-$k$ sampling, as $k$ increases, the probability of sampling a specific token $z_i$ can either decrease (a larger $k$ results in more tokens available for sampling, potentially decreasing the probability of sampling $z_i$) or increase (if at smaller $k$, $z_i$ has zero probability). 
This means $(n, p)$-discoverable extraction rates are not properly ordered according to the choice of $k$, $q$, or $T$, as the underlying token probabilities are affected by these choices.\looseness=-1 

\paragraph{Comparing different schemes (and their cost).} 
Recall that we design our metric for $(n, p)$-dis\-cov\-er\-a\-ble extraction such that the  expected extraction rate matches the amount of memorized content emitted when an end user interacts with the model. 
However, this is challenging, as users are generally free to choose the underlying sampling scheme.
If we report an extraction rate under a choice of temperature $T$ and an end user chooses to use a different temperature $T'\neq T$, is the reported extraction rate still useful? 
In brief, yes, it is still useful, given the trends we observe in Figure~\ref{fig: sampling_schemes}. 
Further, if a practitioner is concerned about the varying extraction rates under different sampling hyperparameters, it is straightforward to compute rates over different choices, as we have done in \Cref{fig: sampling_schemes}.
Because top-$k$, top-$q$, and temperature sampling are post-processing functions applied on top of the generated logit distribution over tokens, it is  cheap to compute these rates over different sampling hyperparameters.

\section{Comparing suffix perplexity scores to their sampling probabilities}\label{app:sec:perplexity}

\begin{figure*}[htbp!]
\noindent \textbf{Row 1: Perplexity-score distributions for target suffixes}
\vspace{.2cm}

\captionsetup[subfigure]{justification=centering}
 \captionsetup[subfigure]{width=\textwidth}
\begin{subfigure}[t]{0.333\textwidth}
\centering
    \includegraphics[width=\linewidth]{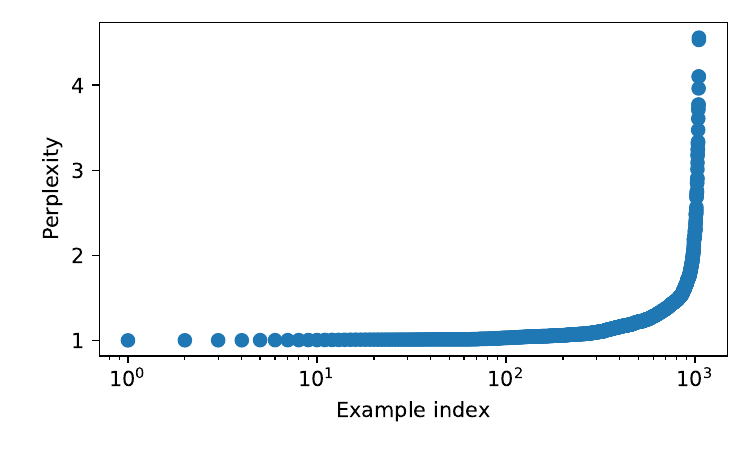}
        \caption{top-$k$, $k=40$, $T=1$}
        \label{fig: ppxtopk}
\end{subfigure}%
\begin{subfigure}[t]{0.333\textwidth}
\centering
    \includegraphics[width=\linewidth]{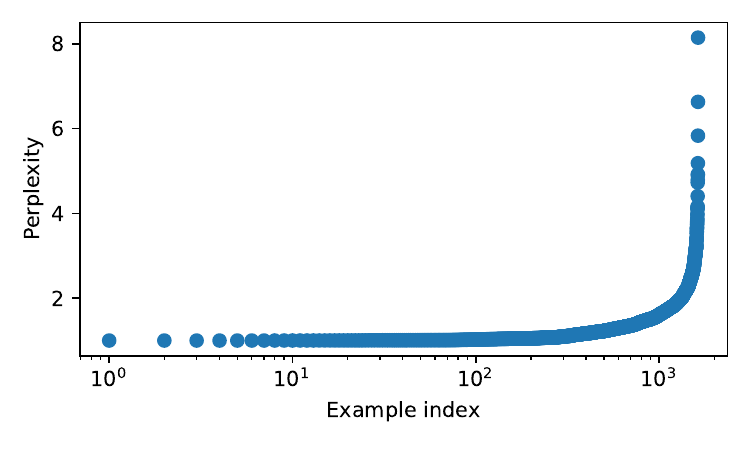}
        \caption{top-$q$, $q=0.9$, $T=1$}
        \label{fig: ppxtopq}
\end{subfigure}%
\begin{subfigure}[t]{0.333\textwidth}
\centering
    \includegraphics[width=\linewidth]{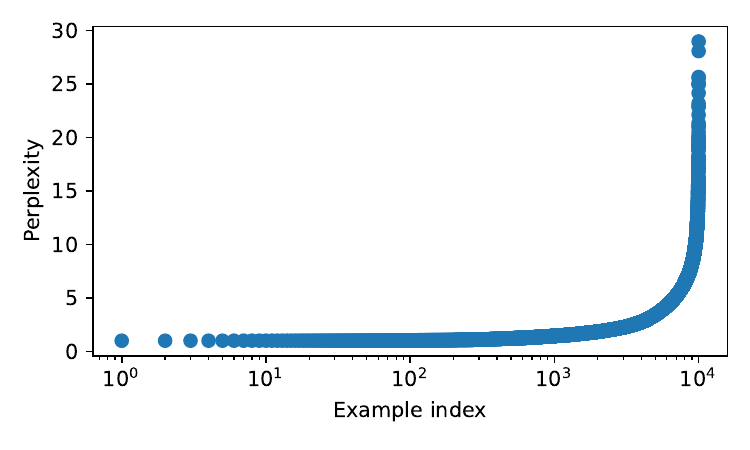}
        \caption{$T=1$}
        \label{fig: ppxt}
\end{subfigure}

\vspace{.4cm}
\noindent \textbf{Row 2: Probability distributions for sampling target suffixes within $n=100$ trials}
\vspace{.2cm}

\begin{subfigure}[t]{0.333\textwidth}
\centering
    \includegraphics[width=\linewidth]{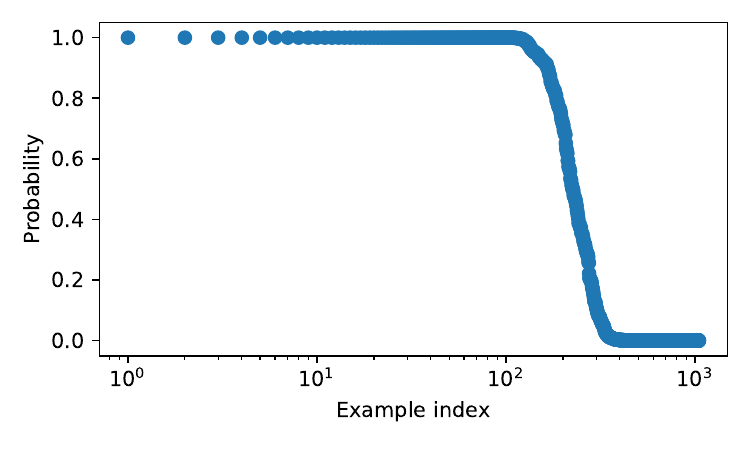}
        \caption{top-$k$, $k=40$, $T=1$}
        \label{fig: probtopk}
\end{subfigure}%
\begin{subfigure}[t]{0.333\textwidth}
\centering
   \includegraphics[width=\linewidth]{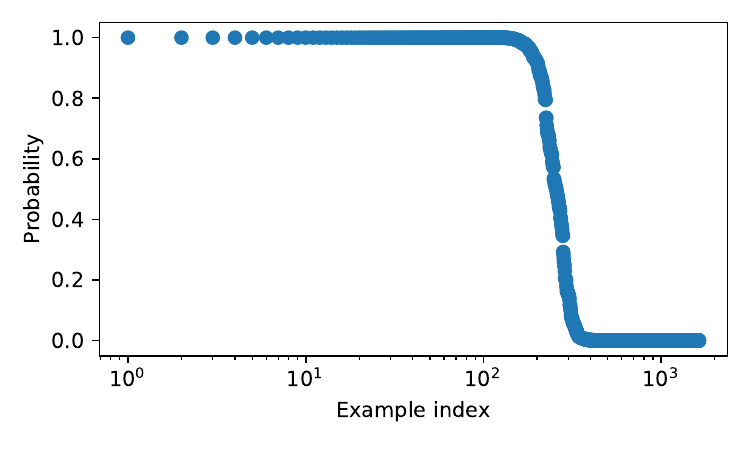}
        \caption{top-$q$, $q=0.9$, $T=1$}
        \label{fig: probtopq}
\end{subfigure}%
\begin{subfigure}[t]{0.333\textwidth}
\centering
    \includegraphics[width=\linewidth]{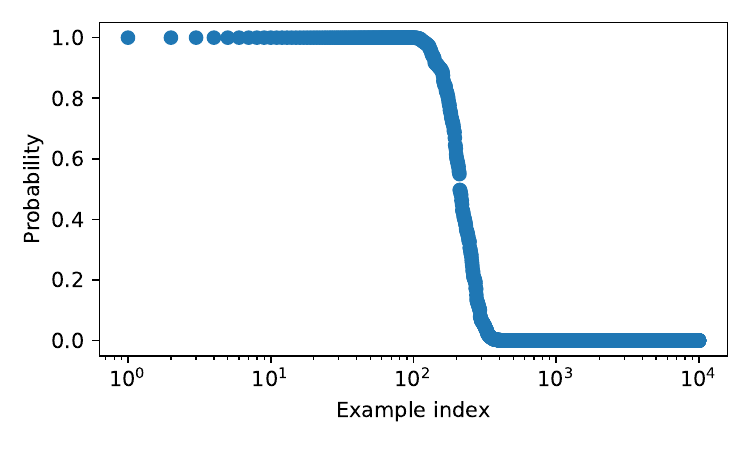}
        \caption{$T=1$}
        \label{fig: probt}
\end{subfigure}
\caption{For each of the $10,000$ Enron examples and Pythia 2.8B, we plot the perplexity score of the target suffix for the given sampling scheme (\textbf{Row 1}). 
We also plot the probability that each of the target suffixes will be sampled within $n=100$ trial (\textbf{Row 2}). 
Note, for top-$k$ and top-$q$ sampling, there are fewer than $10,000$ examples plotted. 
This is because many target suffixes have $0$ probability of being sampled under the sampling scheme.}
\label{fig: ppxandprob}
\end{figure*}

In this appendix, we dig a bit deeper into the probabilities of sampling target suffixes---i.e., extracting targets at generation time. 
For one setting each of top-$k$, top-$q$, and temperature sampling, we plot the distribution over $10,000$ examples in Enron of the probability of sampling the target suffix within $n=100$ trials using Pythia 2.8B (Figures ~\ref{fig: probtopk}, \ref{fig: probtopq}, \& \ref{fig: probt}). 
As a point of comparison, we also plot the perplexity-score distribution for these same examples using the same sampling schemes (Figures ~\ref{fig: ppxtopk}, \ref{fig: ppxtopq}, \& \ref{fig: ppxt}).
That is, in Figure~\ref{fig: ppxandprob}, each column shows the perplexity-score distribution and probability distribution for target suffixes under a different sampling scheme. 

\paragraph{General comments about this visualization.} 
We use top-$k$ with $k=40$ and $T=1$, top-$q$ with $q=0.9$ and $T=1$, and random sampling with $T=1$ in each column, respectively. 
Top-$k$ and top-$q$ sampling both limit the possible tokens that could be generated in each iteration (Section~\ref{sec:prelim:sampling}, Appendix~\ref{sec:sampling_schemes}). 
For both, this means that some suffixes may not be possible generations; 
for the given prefix, the suffix might not be able to be generated because a target token might have zero probability by falling out of the top-$k$-determined or top-$q$-determined choices. 
This is not the case for other sampling schemes, like random sampling with temperature $T$, where every token has nonzero (but possibly low)  probability of being generated at each iteration.

As a result, for both top-$k$ and top-$q$ sampling below (but not random sampling), there are fewer than $10,000$ examples plotted in each distribution.
The top-$k$ probability distribution shows about $1000$ examples and the top-$q$ probability distribution show slightly more than $1000$ examples.
This is because roughly $9000$ suffixes for top-$k$ sampling have zero probability, and similarly for slightly fewer than $9000$ examples for top-$q$ sampling. 
Of course, there will be no perplexity scores for suffixes that the model cannot generate under the given sampling scheme (i.e., those with zero probability). 
So, similarly, the top-$k$ and top-$q$ perplexity distributions show roughly $1000$ and slightly more than $1000$ suffixes, respectively.\looseness=-1

\FloatBarrier

\paragraph{Specific observations about Figure~\ref{fig: ppxandprob}.} 
For top-$k$ sampling, out of the $1000$ suffixes that could be sampled successfully, the majority have extremely low perplexity (\Cref{fig: ppxtopk}). 
These perplexity scores line up with the observation that a large fraction of the $1000$ examples will almost certainly be sampled within $n=100$ trials (\Cref{fig: probtopk}); 
there is not much ``suprise'' (i.e., there is low perplexity) associated with suffixes that are almost guaranteed to be generated. 
A similar set of observations can be made for top-$q=0.9$ sampling in \Cref{fig: ppxtopq,fig: probtopq}.

For random sampling with $T=1$, in theory all possible sequences could potentially be sampled (\Cref{fig: probt}).
As a result, each of the $10,000$ suffixes is reflected in both distributions; each suffix has a defined perplexity score (\Cref{fig: ppxt}). 
Again, we see that a large fraction of target suffixes have small perplexity scores and high probabilities. 
From \Cref{fig: probt}, we see that approximately $250$ target suffixes will almost surely be sampled within $n=100$ trials.

\clearpage
\section{Experimental results over more datasets and model classes}\label{app:sec:moreresults}

In this appendix, we demonstrate that our experimental findings are not constrained to Pythia models evaluated on the Enron dataset, which is the main focus of the results we present in the main paper.
For clarity of presentation, we intentionally discuss results for one model family in the main paper.
Here, we present experimental results over a number of datasets and model classes (Appendix~\ref{app:sec:setup}). 
We show additional results for Pythia (\Cref{app:sec:more:pythia}), as well as results on Llama (\Cref{app:sec:more:llama}) and OPT (\Cref{app:sec:more:opt}). 

\subsection{Pythia model family}\label{app:sec:more:pythia}


\begin{figure*}[htbp!]
\begin{minipage}{0.045\textwidth}
    \textbf{\;}
    \vspace{.3cm}
\end{minipage}%
\hfill
\begin{minipage}{0.312\textwidth}
    \centering
    \textbf{\hspace{.6cm}Top-$k$ ($k=40$, $T=1$)}
    \vspace{.3cm}
\end{minipage}%
\hfill
\begin{minipage}{0.312\textwidth}
    \centering
    \textbf{\hspace{.6cm}Top-$q$ ($q=0.9$, $T=1$)}
    \vspace{.3cm}
\end{minipage}%
\hfill
\begin{minipage}{0.312\textwidth}
    \centering
    \textbf{\hspace{.6cm}Random ($T=1$)}
    \vspace{.3cm}
\end{minipage}
\captionsetup[subfigure]{justification=centering}
  \centering
\begin{subfigure}[t]{0.045\textwidth}
\vspace{-1.65cm}
\textbf{1B}
\vspace{1cm}
\end{subfigure}%
\hfill
\begin{subfigure}[t]{0.312\textwidth}
\centering
    \includegraphics[width=1.\linewidth]{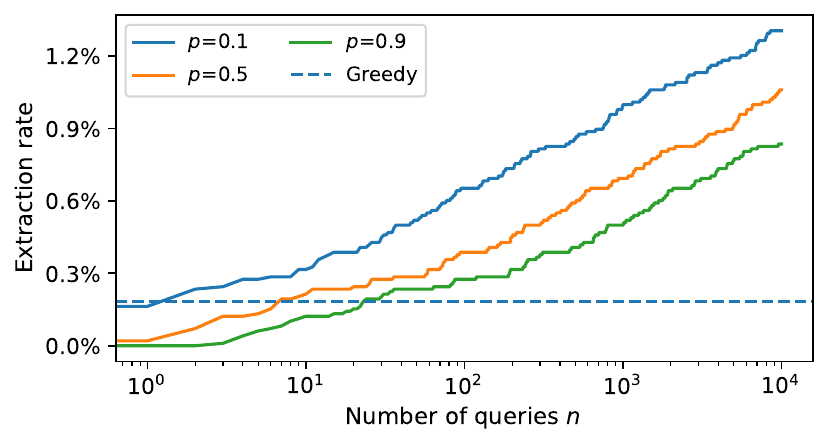}
        \caption{Pythia 1B, top-$k$}
        \vspace{.5cm}
        \label{fig:pythia-wiki-1b-topk}
\end{subfigure}%
\hfill
\begin{subfigure}[t]{0.312\textwidth}
\centering
    \includegraphics[width=1.\linewidth]{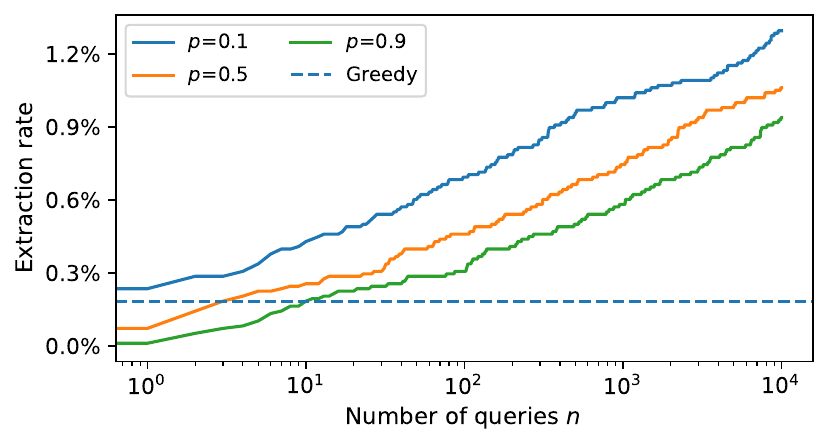}
        \caption{Pythia 1B, top-$q$}
        \vspace{.5cm}
        \label{fig:pythia-wiki-1b-topq}
\end{subfigure}%
\hfill
\begin{subfigure}[t]{0.312\textwidth}
\centering
     \includegraphics[width=1.\linewidth]{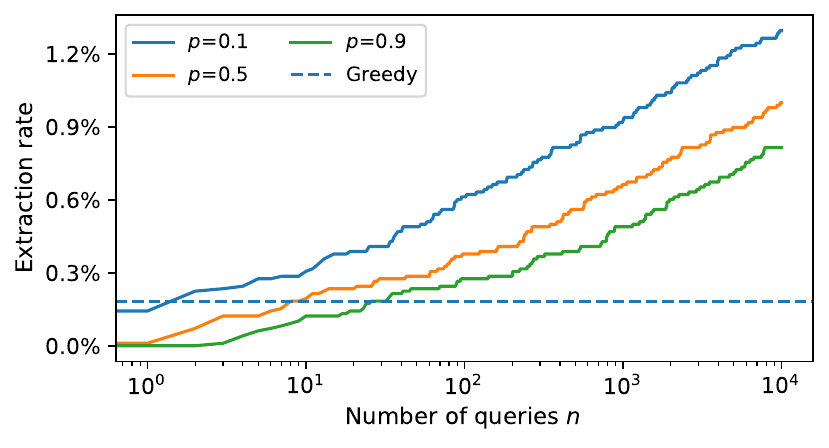}
        \caption{Pythia 1B, $T=1$}
        \vspace{.5cm}
        \label{fig:pythia-wiki-1b-random}
\end{subfigure}\\%
\begin{subfigure}[t]{0.045\textwidth}
\vspace{-1.65cm}
\textbf{2.8B}
\vspace{1cm}
\end{subfigure}%
\hfill
\captionsetup[subfigure]{justification=centering}
  \centering
\begin{subfigure}[t]{0.312\textwidth}
\centering
    \includegraphics[width=1.\linewidth]{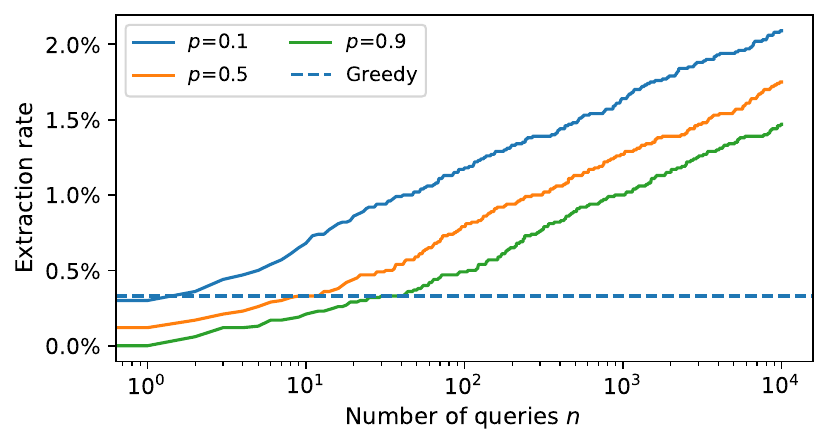}
        \caption{Pythia 2.8B, top-$k$}
        \vspace{.5cm}
        \label{fig:pythia-wiki-2.8b-topk}
\end{subfigure}%
\hfill
\begin{subfigure}[t]{0.312\textwidth}
\centering
    \includegraphics[width=1.\linewidth]{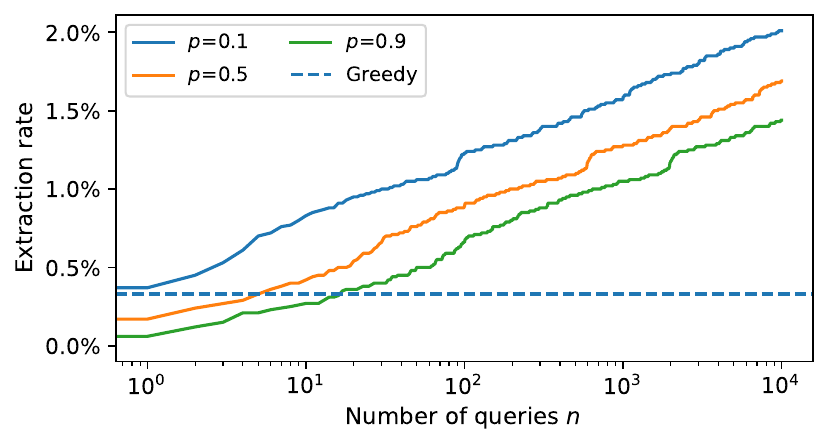}
        \caption{Pythia 2.8B, top-$q$}
        \vspace{.5cm}
        \label{fig:pythia-wiki-2.8b-topq}
\end{subfigure}
\hfill
\begin{subfigure}[t]{0.312\textwidth}
\centering
         \includegraphics[width=1.\linewidth]{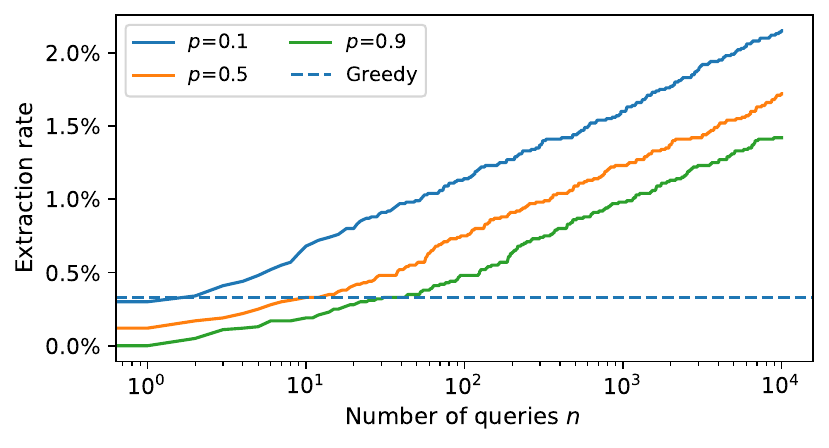}
        \caption{Pythia 2.8B, $T=1$}
        \vspace{.5cm}
        \label{fig:pythia-wiki-2.8b-random}
\end{subfigure}\\
\begin{subfigure}[t]{0.045\textwidth}
\vspace{-1.65cm}
\textbf{6.9B}
\vspace{1cm}
\end{subfigure}%
\hfill
\begin{subfigure}[t]{0.312\textwidth}
\centering
    \includegraphics[width=1.\linewidth]{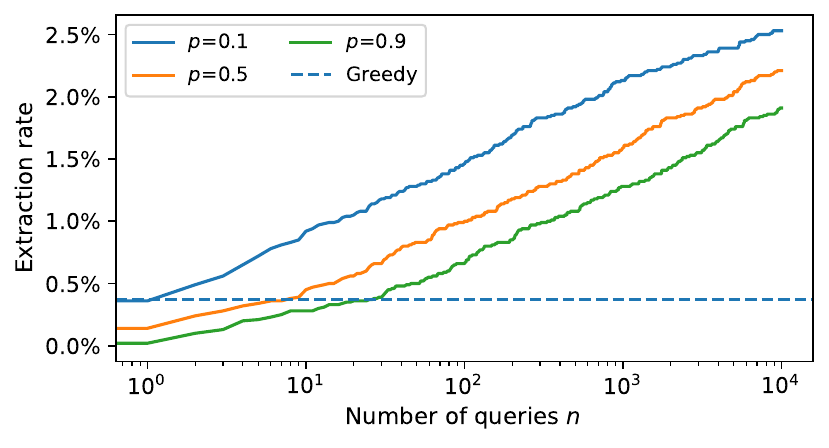}
        \caption{Pythia 6.9B, top-$k$}
        \vspace{.5cm}
        \label{fig:pythia-wiki-6.9b-topk}
\end{subfigure}%
\hfill
\begin{subfigure}[t]{0.312\textwidth}
\centering
    \includegraphics[width=1.\linewidth]{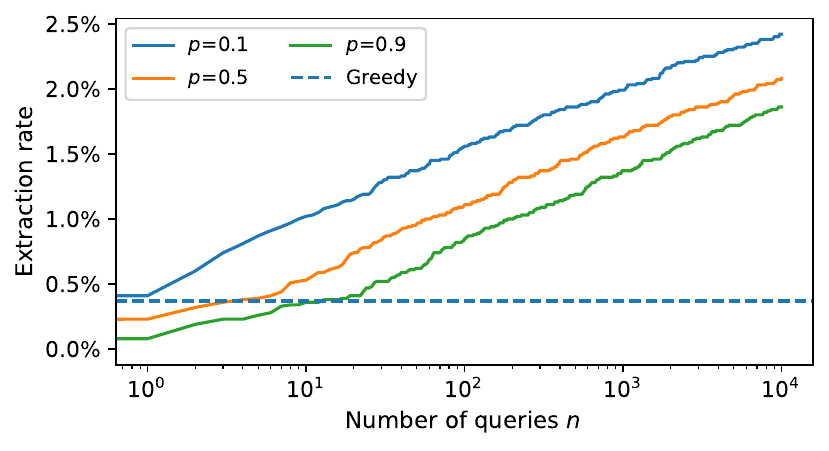}
        \caption{Pythia 6.9B, top-$q$}
        \vspace{.5cm}
        \label{fig:pythia-wiki-6.9b-topq}
\end{subfigure}
\hfill
\begin{subfigure}[t]{0.312\textwidth}
\centering
         \includegraphics[width=1.\linewidth]{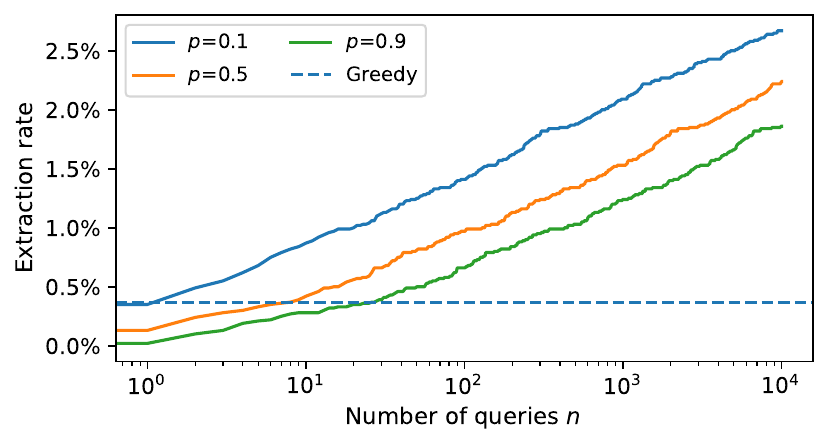}
        \caption{Pythia 6.9B, $T=1$}
        \vspace{.5cm}
        \label{fig:pythia-wiki-6.9b-random}
\end{subfigure}\\%
\begin{subfigure}[t]{0.045\textwidth}
\vspace{-1.65cm}
\textbf{12B}
\vspace{1cm}
\end{subfigure}%
\hfill
\begin{subfigure}[t]{0.312\textwidth}
\centering
    \includegraphics[width=1.\linewidth]{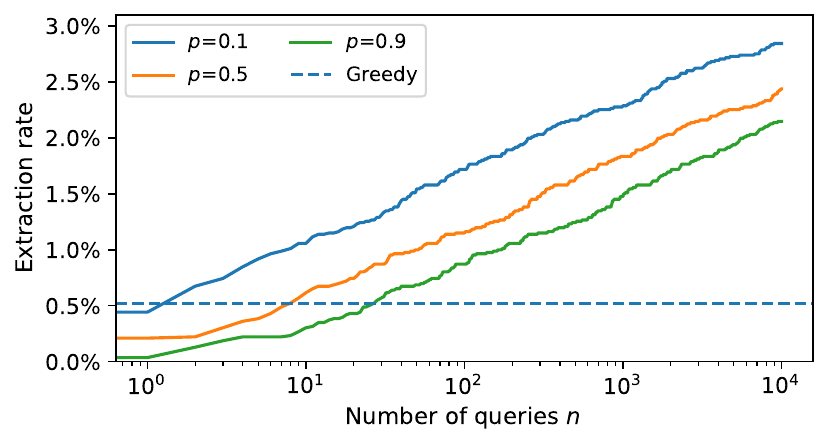}
        \caption{Pythia 12B, top-$k$}
        \vspace{.5cm}
        \label{fig:pythia-wiki-12b-topk}
\end{subfigure}%
\hfill
\begin{subfigure}[t]{0.312\textwidth}
\centering
    \includegraphics[width=1.\linewidth]{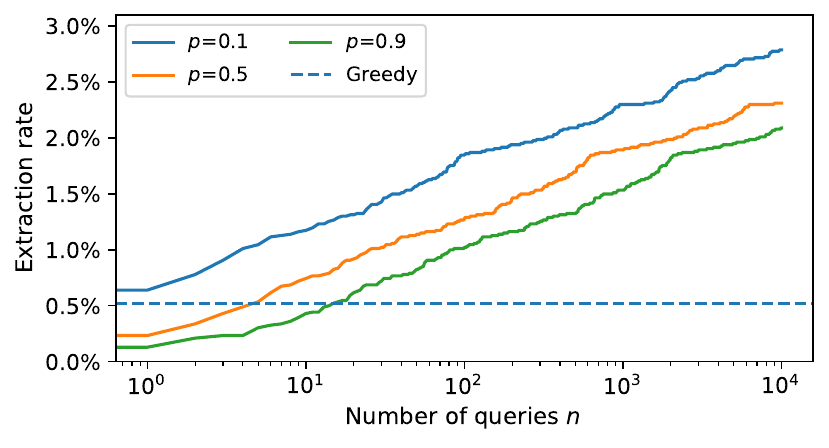}
        \caption{Pythia 12B, top-$q$}
        \vspace{.5cm}
        \label{fig:pythia-wiki-12b-topq}
\end{subfigure}
\hfill
\begin{subfigure}[t]{0.312\textwidth}
\centering
         \includegraphics[width=1.\linewidth]{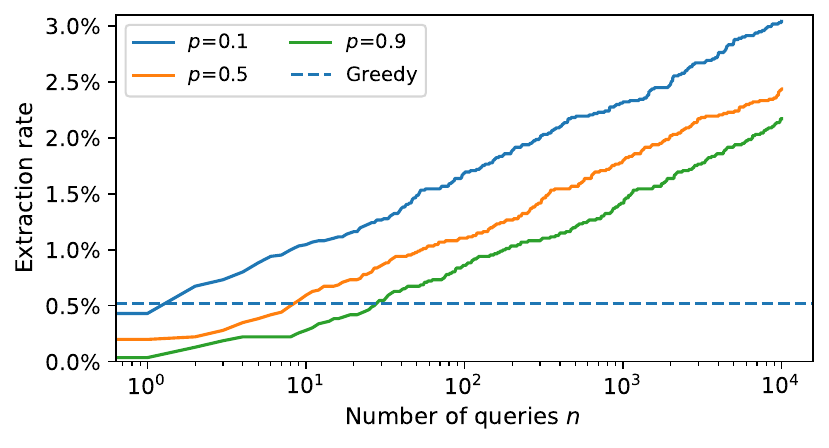}
        \caption{Pythia 12B, $T=1$}
        \vspace{.5cm}
        \label{fig:pythia-wiki-12b-random}
\end{subfigure}\\
\caption{\textbf{Pythia models on The Pile's Wikipedia subset.} Comparison of greedy-sampled discoverable extraction rates and $(n, p)$-discoverable extraction rates for different-sized Pythia models (1B, 2.8B, 6.9B, and 12B) for different sampling schemes (top-$k$ with $k=40$ and $T=1$; top-$q$ with $q=0.1$ and $T=1$; random with $T=1$) for Wikipedia ($10,000$ examples). 
Each row is for a different model, and each column is for a different sampling scheme.
For the same model, extraction-rate curves at different values of $p$ are fairly consistent across sampling schemes.
Larger models exhibit higher extraction rates.
Across all models sizes, for the same settings of $n$ and $p$, the extraction rates are lower than for Pythia on Enron (Figures~\ref{fig: app_model_sizes}~\&~\ref{fig: sampling_schemes}) and on GitHub (Figure~\ref{fig:pythia-github}).}
\label{fig:pythia-wiki}
\end{figure*}
\FloatBarrier


\clearpage
\begin{figure*}[htbp!]
\begin{minipage}{0.045\textwidth}
    \textbf{\;}
    \vspace{.3cm}
\end{minipage}%
\hfill
\begin{minipage}{0.312\textwidth}
    \centering
    \textbf{\hspace{.6cm}Top-$k$ ($k=40$, $T=1$)}
    \vspace{.3cm}
\end{minipage}%
\hfill
\begin{minipage}{0.312\textwidth}
    \centering
    \textbf{\hspace{.6cm}Top-$q$ ($q=0.9$, $T=1$)}
    \vspace{.3cm}
\end{minipage}%
\hfill
\begin{minipage}{0.312\textwidth}
    \centering
    \textbf{\hspace{.6cm}Random ($T=1$)}
    \vspace{.3cm}
\end{minipage}
\captionsetup[subfigure]{justification=centering}
  \centering
\begin{subfigure}[t]{0.045\textwidth}
\vspace{-1.65cm}
\textbf{1B}
\vspace{1cm}
\end{subfigure}%
\hfill
\begin{subfigure}[t]{0.312\textwidth}
\centering
    \includegraphics[width=1.\linewidth]{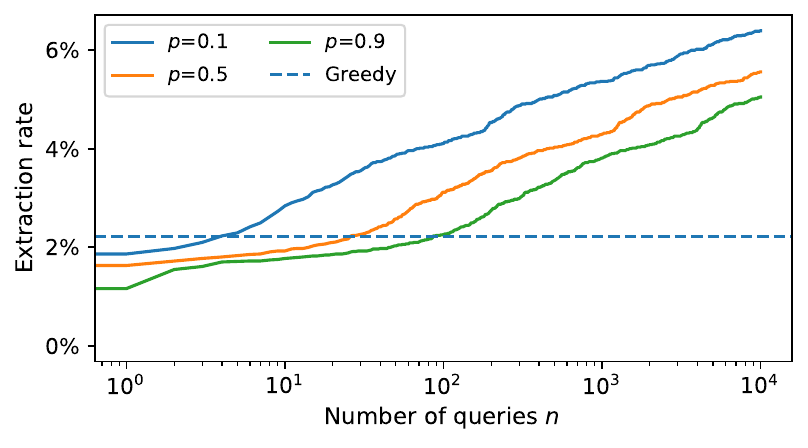}
        \caption{Pythia 1B, top-$k$}
        \vspace{.5cm}
        \label{fig:pythia-github-1b-topk}
\end{subfigure}%
\hfill
\begin{subfigure}[t]{0.312\textwidth}
\centering
    \includegraphics[width=1.\linewidth]{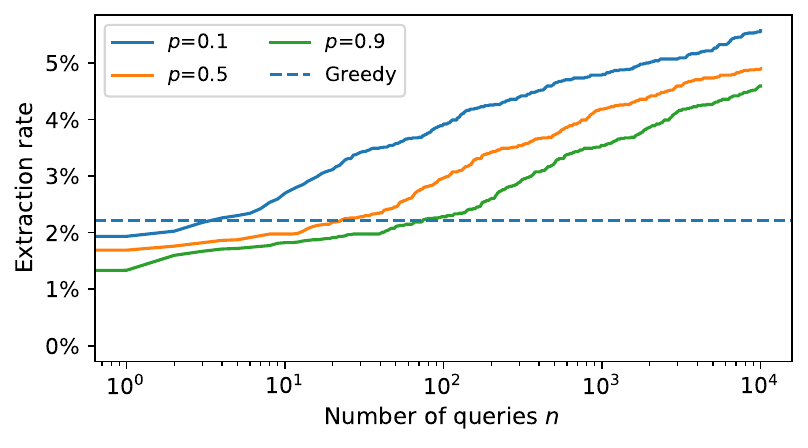}
        \caption{Pythia 1B, top-$q$}
        \vspace{.5cm}
        \label{fig:pythia-github-1b-topq}
\end{subfigure}%
\hfill
\begin{subfigure}[t]{0.312\textwidth}
\centering
    \includegraphics[width=1.\linewidth]{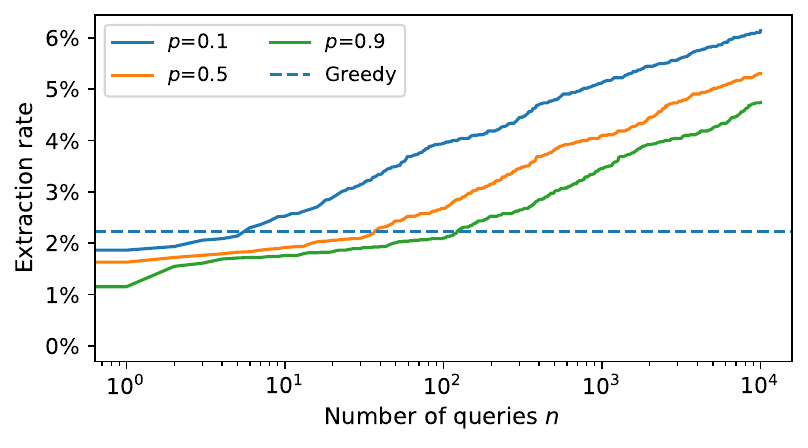}
        \caption{Pythia 1B, $T=1$}
        \vspace{.5cm}
        \label{fig:pythia-github-1b-random}
\end{subfigure}\\%
\begin{subfigure}[t]{0.045\textwidth}
\vspace{-1.65cm}
\textbf{2.8B}
\vspace{1cm}
\end{subfigure}%
\hfill
\captionsetup[subfigure]{justification=centering}
  \centering
\begin{subfigure}[t]{0.312\textwidth}
\centering
\includegraphics[width=1.\linewidth]{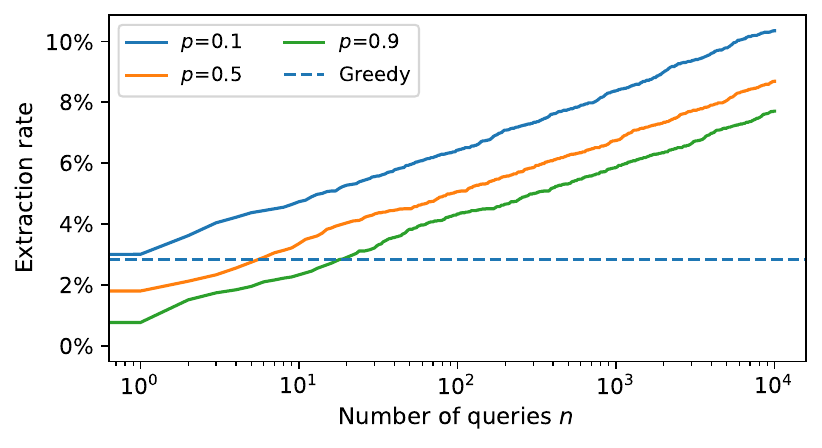}
        \caption{Pythia 2.8B, top-$k$}
        \vspace{.5cm}
        \label{fig:pythia-github-2.8b-topk}
\end{subfigure}%
\hfill
\begin{subfigure}[t]{0.312\textwidth}
\centering
\includegraphics[width=1.\linewidth]{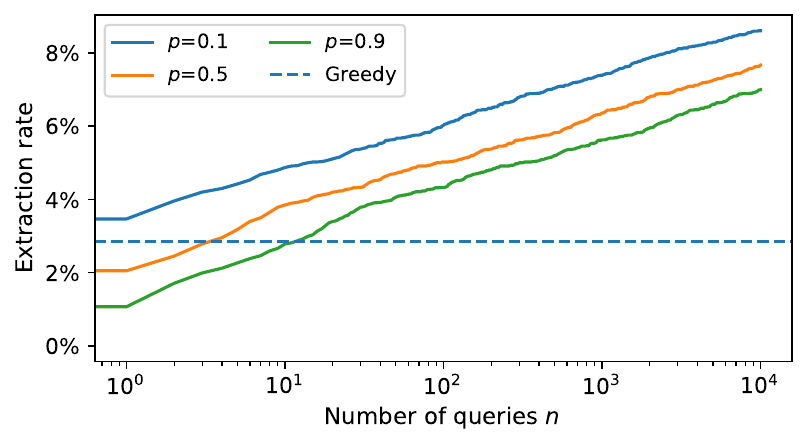}
        \caption{Pythia 2.8B, top-$q$}
        \vspace{.5cm}
        \label{fig:pythia-github-2.8b-topq}
\end{subfigure}
\hfill
\begin{subfigure}[t]{0.312\textwidth}
\centering
\includegraphics[width=1.\linewidth]{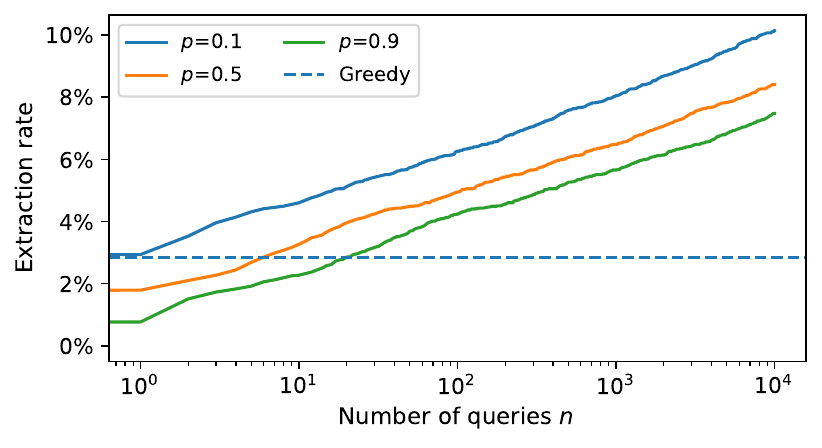}
        \caption{Pythia 2.8B, $T=1$}
        \vspace{.5cm}
        \label{fig:pythia-github-2.8b-random}
\end{subfigure}\\
\begin{subfigure}[t]{0.045\textwidth}
\vspace{-1.65cm}
\textbf{6.9B}
\vspace{1cm}
\end{subfigure}%
\hfill
\begin{subfigure}[t]{0.312\textwidth}
\centering
    \includegraphics[width=1.\linewidth]{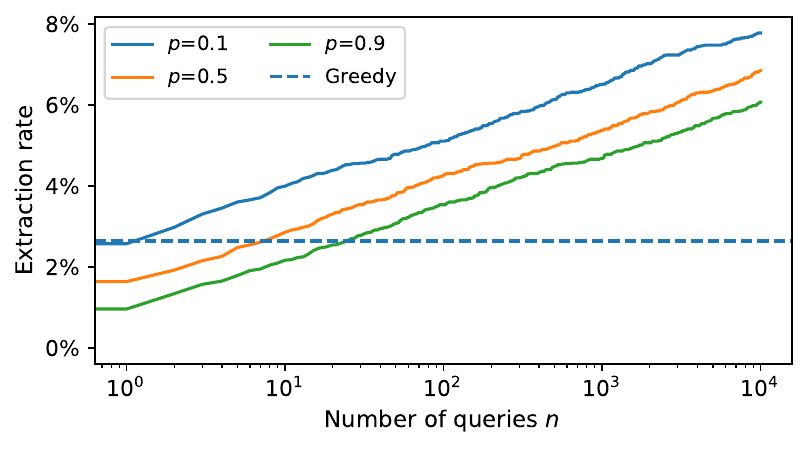}
        \caption{Pythia 6.9B, top-$k$}
        \vspace{.5cm}
        \label{fig:pythia-github-6.9b-topk}
\end{subfigure}%
\hfill
\begin{subfigure}[t]{0.312\textwidth}
\centering
    \includegraphics[width=1.\linewidth]{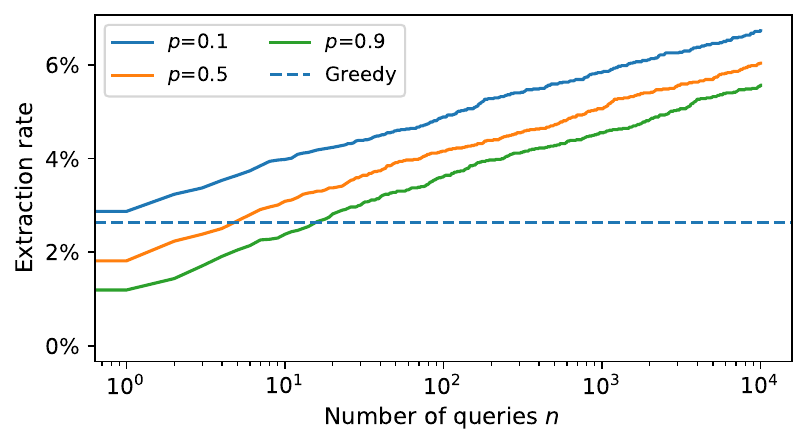}
        \caption{Pythia 6.9B, top-$q$}
        \vspace{.5cm}
        \label{fig:pythia-github-6.9b-topq}
\end{subfigure}
\hfill
\begin{subfigure}[t]{0.312\textwidth}
\centering
    \includegraphics[width=1.\linewidth]{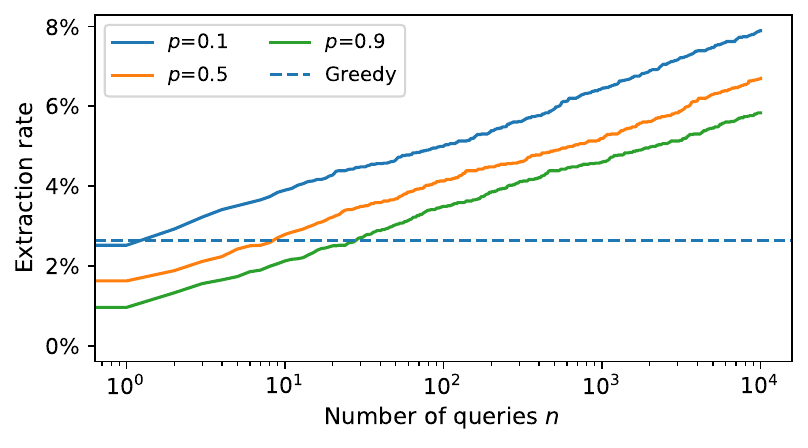}
        \caption{Pythia 6.9B, $T=1$}
        \vspace{.5cm}
        \label{fig:pythia-github-6.9b-random}
\end{subfigure}\\%
\begin{subfigure}[t]{0.045\textwidth}
\vspace{-1.65cm}
\textbf{12B}
\vspace{1cm}
\end{subfigure}%
\hfill
\begin{subfigure}[t]{0.312\textwidth}
\centering
    \includegraphics[width=1.\linewidth]{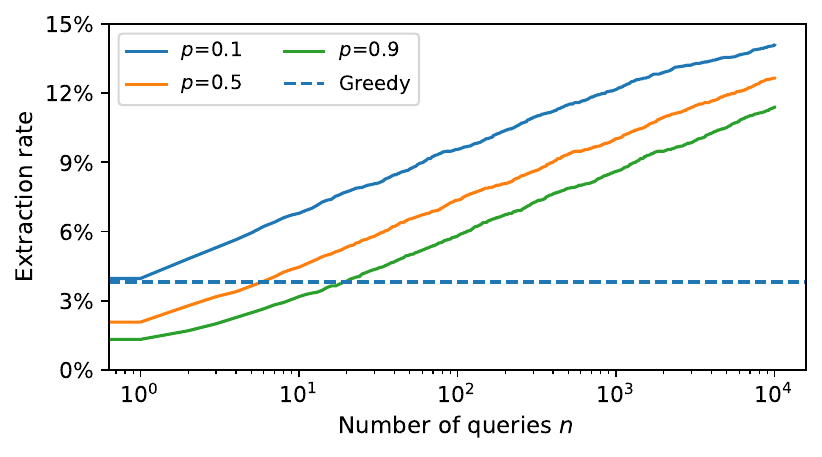}
        \caption{Pythia 12B, top-$k$}
        \vspace{.5cm}
        \label{fig:pythia-github-12b-topk}
\end{subfigure}%
\hfill
\begin{subfigure}[t]{0.312\textwidth}
\centering
    \includegraphics[width=1.\linewidth]{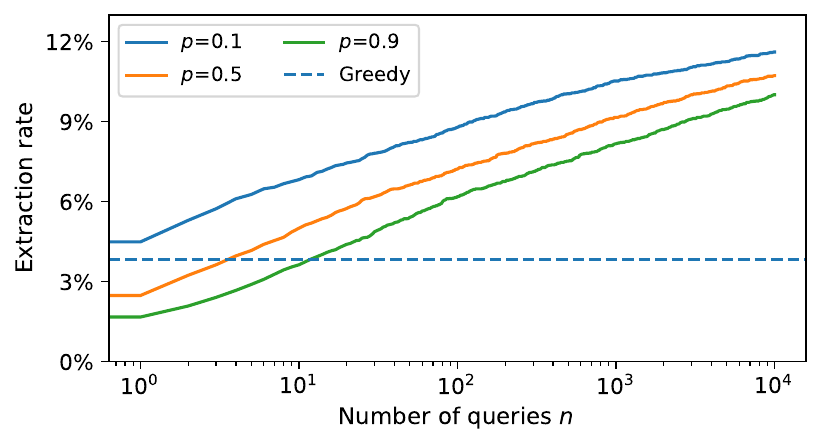}
        \caption{Pythia 12B, top-$q$}
        \vspace{.5cm}
        \label{fig:pythia-github-12b-topq}
\end{subfigure}
\hfill
\begin{subfigure}[t]{0.312\textwidth}
\centering
    \includegraphics[width=1.\linewidth]{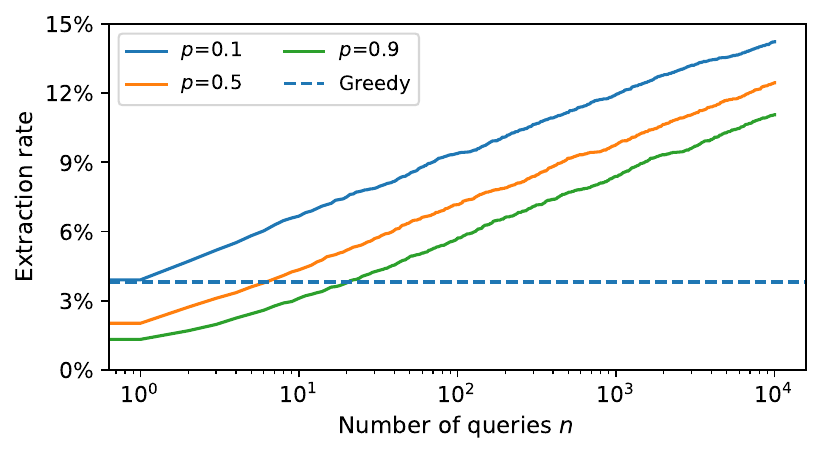}
        \caption{Pythia 12B, $T=1$}
        \vspace{.5cm}
        \label{fig:pythia-github-12b-random}
\end{subfigure}\\
\caption{\textbf{Pythia models on The Pile's GitHub subset.} 
Comparison of greedy-sampled discoverable extraction rates and $(n, p)$-discoverable extraction rates for different-sized Pythia models (1B, 2.8B, 6.9B, and 12B) for different sampling schemes (top-$k$ with $k=40$ and $T=1$; top-$q$ with $q=0.1$ and $T=1$; random with $T=1$) for GitHub ($10,000$ examples). 
Each row is for a different model, and each column is for a different sampling scheme.
For the same model, extraction-rate curves at different values of $p$ are fairly consistent across top-$k$ and random $T=1$ sampling; top-$q$ extraction rates are slightly lower.
(This is a similar pattern as for Llama on Common Crawl and OPT on Common Crawl; see Figure~\ref{fig:llama-cc} and Figure~\ref{fig:opt-cc}, respectively.) 
Larger models tend to exhibit higher extraction rates for the plotted $n$ and $p$. 
One exception is Pythia 6.9B, which (for the given $n$ and $p$) exhibits lower extraction rates than Pythia 2.8B;
Pythia 12B exhibits the highest extraction rate. 
Across all models sizes, for the same settings of $n$ and $p$, the extraction rates are higher than for Pythia on Enron (Figures~\ref{fig: app_model_sizes}~\&~\ref{fig: sampling_schemes}) and for Pythia on Wikipedia (Figure~\ref{fig:pythia-wiki}).}
\label{fig:pythia-github}
\end{figure*}
\FloatBarrier
\clearpage
\subsection{Llama model family}\label{app:sec:more:llama}


\begin{figure*}[htbp!]
\begin{minipage}{0.045\textwidth}
    \textbf{\;}
    \vspace{.3cm}
\end{minipage}%
\hfill
\begin{minipage}{0.312\textwidth}
    \centering
    \textbf{\hspace{.6cm}Top-$k$ ($k=40$, $T=1$)}
    \vspace{.3cm}
\end{minipage}%
\hfill
\begin{minipage}{0.312\textwidth}
    \centering
    \textbf{\hspace{.6cm}Top-$q$ ($q=0.9$, $T=1$)}
    \vspace{.3cm}
\end{minipage}%
\hfill
\begin{minipage}{0.312\textwidth}
    \centering
    \textbf{\hspace{.6cm}Random ($T=1$)}
    \vspace{.3cm}
\end{minipage}
\captionsetup[subfigure]{justification=centering}
  \centering
\begin{subfigure}[t]{0.045\textwidth}
\vspace{-1.65cm}
\textbf{7B}
\vspace{1cm}
\end{subfigure}%
\hfill
\begin{subfigure}[t]{0.312\textwidth}
\centering
    \includegraphics[width=1.\linewidth]{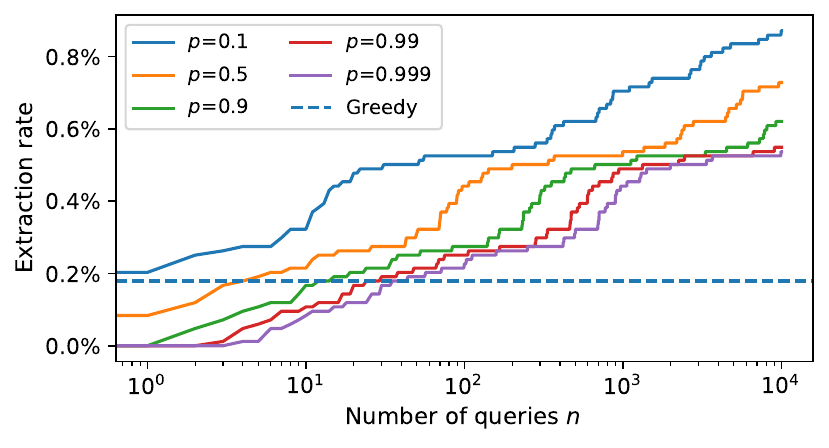}
        \caption{Llama 7B, top-$k$}
        \vspace{.5cm}
        \label{fig:llama-cc-7b-topk}
\end{subfigure}%
\hfill
\begin{subfigure}[t]{0.312\textwidth}
\centering
    \includegraphics[width=1.\linewidth]{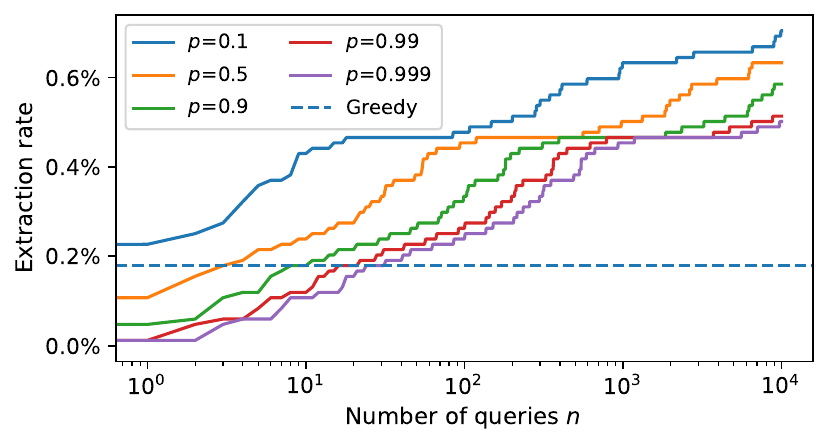}
        \caption{Llama 7B, top-$q$}
        \vspace{.5cm}
        \label{fig:llama-cc-7b-topq}
\end{subfigure}
\hfill
\begin{subfigure}[t]{0.312\textwidth}
\centering
    \includegraphics[width=1.\linewidth]{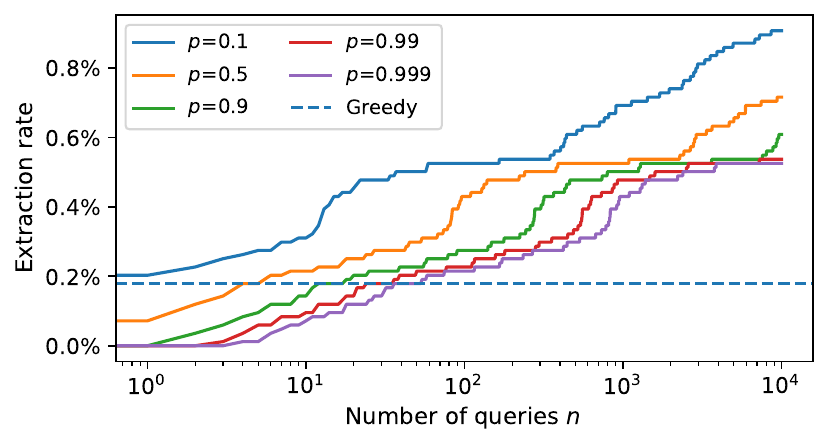}
        \caption{Llama 7B, $T=1$}
        \vspace{.5cm}
        \label{fig:llama-cc-7b-random}
\end{subfigure}\\%
\begin{subfigure}[t]{0.045\textwidth}
\vspace{-1.65cm}
\textbf{13B}
\vspace{1cm}
\end{subfigure}%
\hfill
\begin{subfigure}[t]{0.312\textwidth}
\centering
    \includegraphics[width=1.\linewidth]{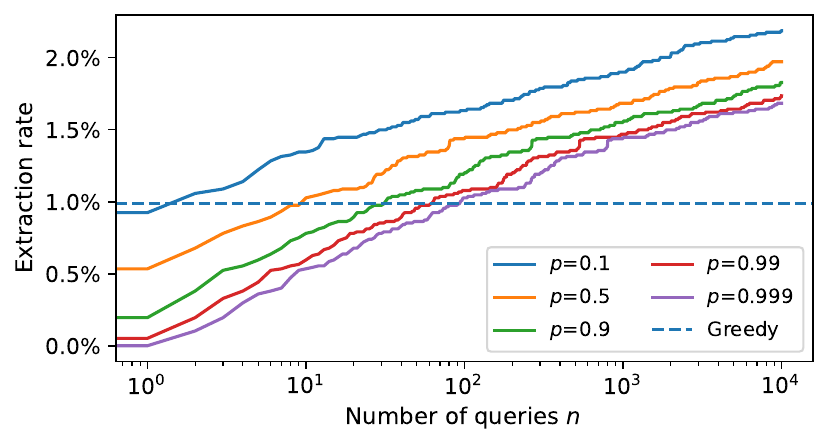}
        \caption{Llama 13B, top-$k$}
        \vspace{.5cm}
        \label{fig:llama-cc-13b-topk}
\end{subfigure}%
\hfill
\begin{subfigure}[t]{0.312\textwidth}
\centering
    \includegraphics[width=1.\linewidth]{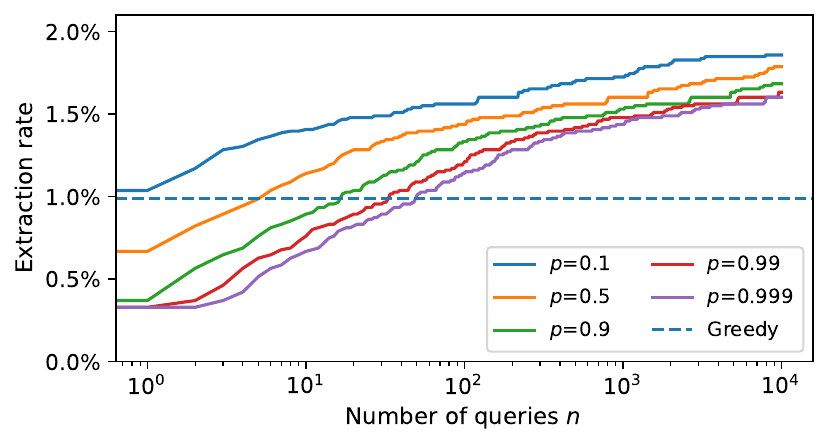}
        \caption{Llama 13B, top-$q$}
        \vspace{.5cm}
        \label{fig:llama-cc-13b-topq}
\end{subfigure}
\hfill
\begin{subfigure}[t]{0.312\textwidth}
\centering
    \includegraphics[width=1.\linewidth]{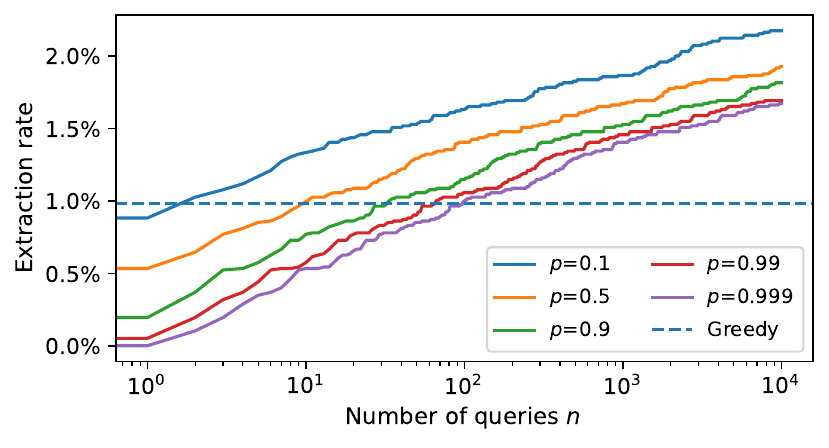}
        \caption{Llama 13B, $T=1$}
        \vspace{.5cm}
        \label{fig:llama-cc-13b-random}
\end{subfigure}\\
\caption{\textbf{Llama models on Common Crawl.} 
Comparison of greedy-sampled discoverable extraction rates and $(n, p)$-discoverable extraction rates for different-sized Llama models (7B and 13B) for different sampling schemes (top-$k$ with $k=40$ and $T=1$; top-$q$ with $q=0.1$ and $T=1$; random with $T=1$) for Common Crawl ($10,000$ examples). 
Each row is for a different model, and each column is for a different sampling scheme.
For the same model, extraction-rate curves at different values of $p$ are fairly consistent across top-$k$ and random $T=1$ sampling; top-$q$ extraction rates are slightly lower.
(This is a similar pattern as for Pythia on GitHub and OPT on Common Crawl; see Figure~\ref{fig:pythia-github} and Figure~\ref{fig:opt-cc}, respectively.) 
The larger Llama 13B model exhibits higher extraction rates than Llama 7B. 
The extraction rates for Llama on Common Crawl are lower than for models of comparable sizes in the Pythia family on all data subsets (Figures~\ref{fig: app_model_sizes}, \ref{fig: sampling_schemes}, \ref{fig:pythia-wiki} \&  \ref{fig:pythia-github}).}
\label{fig:llama-cc}
\end{figure*}
\FloatBarrier
\clearpage
\subsection{OPT model family}\label{app:sec:more:opt}


\begin{figure*}[htbp!]
\begin{minipage}{0.045\textwidth}
    \textbf{\;}
    \vspace{.3cm}
\end{minipage}%
\hfill
\begin{minipage}{0.312\textwidth}
    \centering
    \textbf{\hspace{.6cm}Top-$k$ ($k=40$, $T=1$)}
    \vspace{.3cm}
\end{minipage}%
\hfill
\begin{minipage}{0.312\textwidth}
    \centering
    \textbf{\hspace{.6cm}Top-$q$ ($q=0.9$, $T=1$)}
    \vspace{.3cm}
\end{minipage}%
\hfill
\begin{minipage}{0.312\textwidth}
    \centering
    \textbf{\hspace{.6cm}Random ($T=1$)}
    \vspace{.3cm}
\end{minipage}
\captionsetup[subfigure]{justification=centering}
  \centering
\begin{subfigure}[t]{0.045\textwidth}
\vspace{-1.65cm}
\textbf{\small{350M}}
\vspace{1cm}
\end{subfigure}%
\hfill
\begin{subfigure}[t]{0.312\textwidth}
\centering
    \includegraphics[width=1.\linewidth]{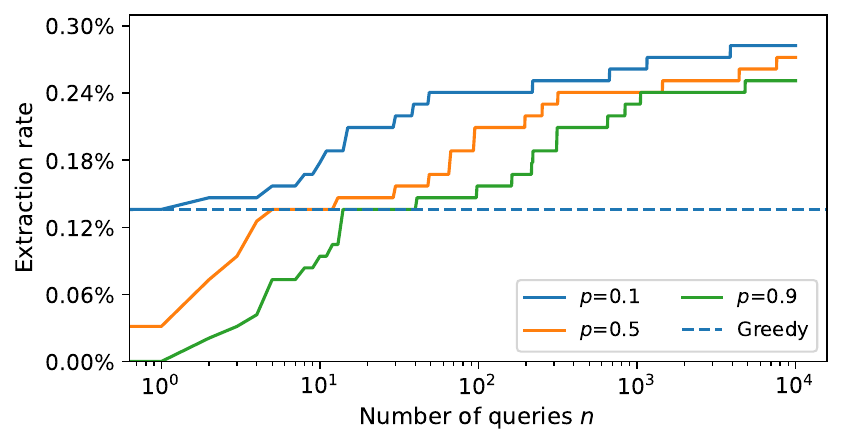}
        \caption{OPT 350M, top-$k$}
        \vspace{.5cm}
        \label{fig:opt-cc-350m-topk}
\end{subfigure}%
\hfill
\begin{subfigure}[t]{0.312\textwidth}
\centering
    \includegraphics[width=1.\linewidth]{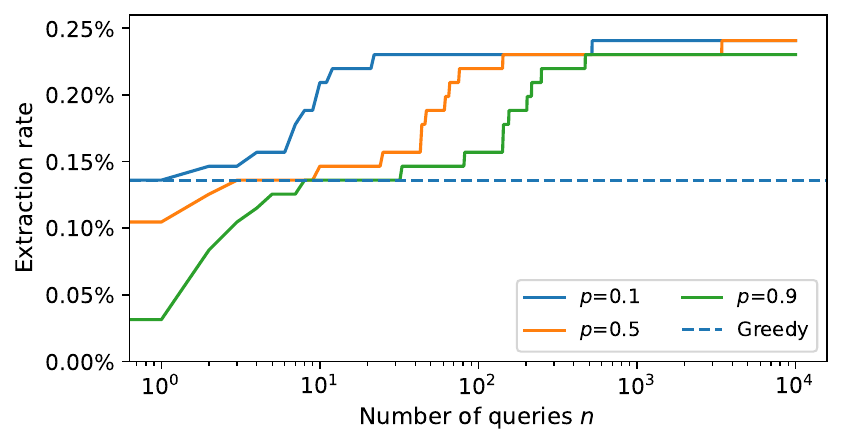}
        \caption{OPT 350M, top-$q$}
        \vspace{.5cm}
        \label{fig:opt-cc-350m-topq}
\end{subfigure}%
\hfill
\begin{subfigure}[t]{0.312\textwidth}
\centering
    \includegraphics[width=1.\linewidth]{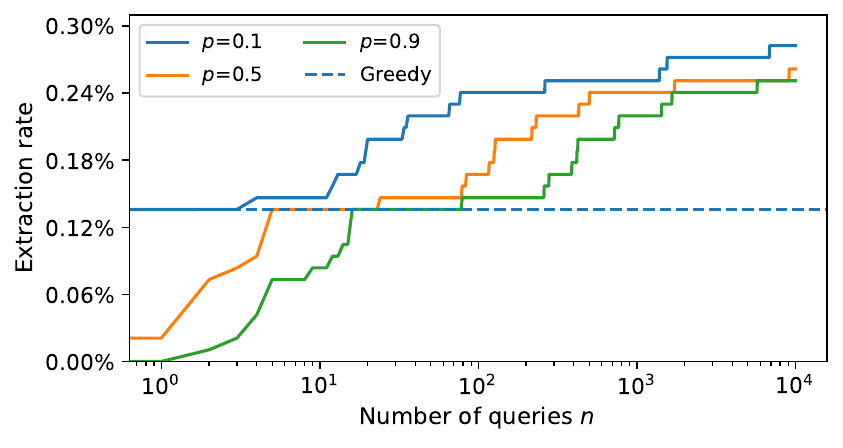}
        \caption{OPT 350M, $T=1$}
        \vspace{.5cm}
        \label{fig:opt-cc-350m-random}
\end{subfigure}\\%
\begin{subfigure}[t]{0.045\textwidth}
\vspace{-1.65cm}
\textbf{1.3B}
\vspace{1cm}
\end{subfigure}%
\hfill
\captionsetup[subfigure]{justification=centering}
  \centering
\begin{subfigure}[t]{0.312\textwidth}
\centering
    \includegraphics[width=1.\linewidth]{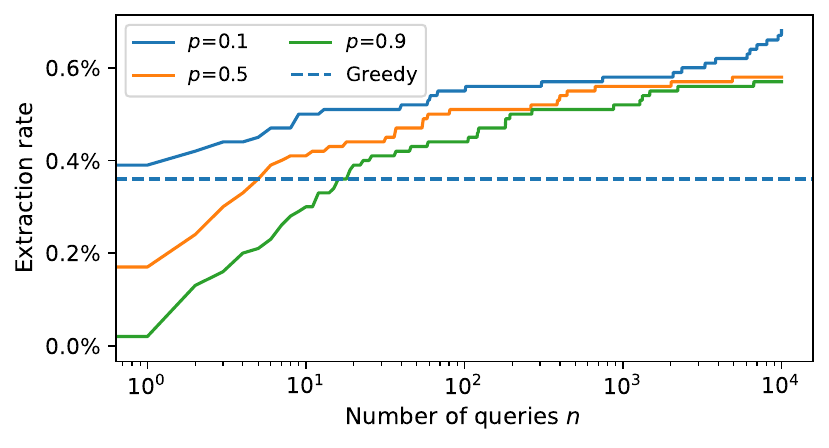}
        \caption{OPT 1.3B, top-$k$}
        \vspace{.5cm}
        \label{fig:opt-cc-1.3b-topk}
\end{subfigure}%
\hfill
\begin{subfigure}[t]{0.312\textwidth}
\centering
    \includegraphics[width=1.\linewidth]{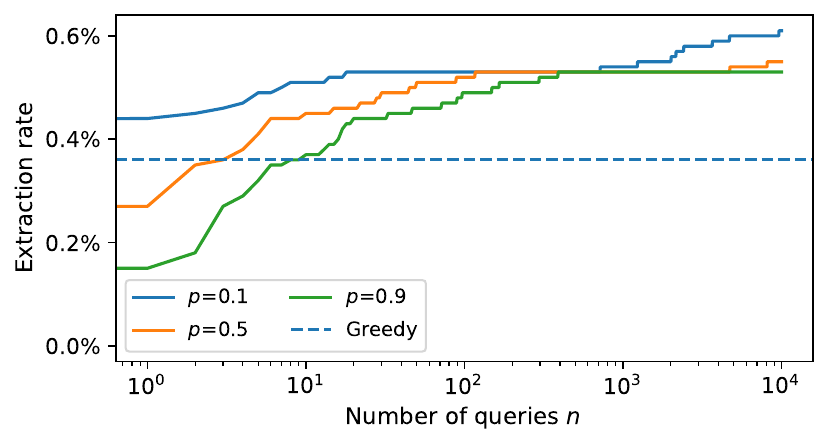}
        \caption{OPT 1.3B, top-$q$}
        \vspace{.5cm}
        \label{fig:opt-cc-1.3b-topq}
\end{subfigure}
\hfill
\begin{subfigure}[t]{0.312\textwidth}
\centering
    \includegraphics[width=1.\linewidth]{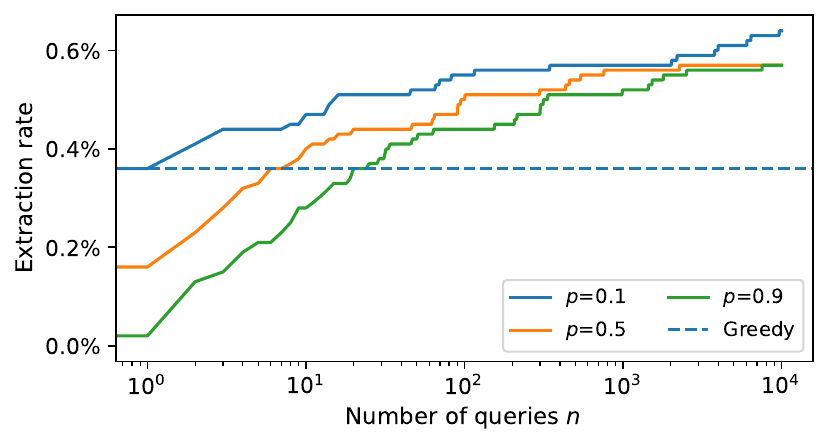}
        \caption{OPT 1.3B, $T=1$}
        \vspace{.5cm}
        \label{fig:opt-cc-1.3b-random}
\end{subfigure}\\
\begin{subfigure}[t]{0.045\textwidth}
\vspace{-1.65cm}
\textbf{2.7B}
\vspace{1cm}
\end{subfigure}%
\hfill
\begin{subfigure}[t]{0.312\textwidth}
\centering
        \includegraphics[width=1.\linewidth]{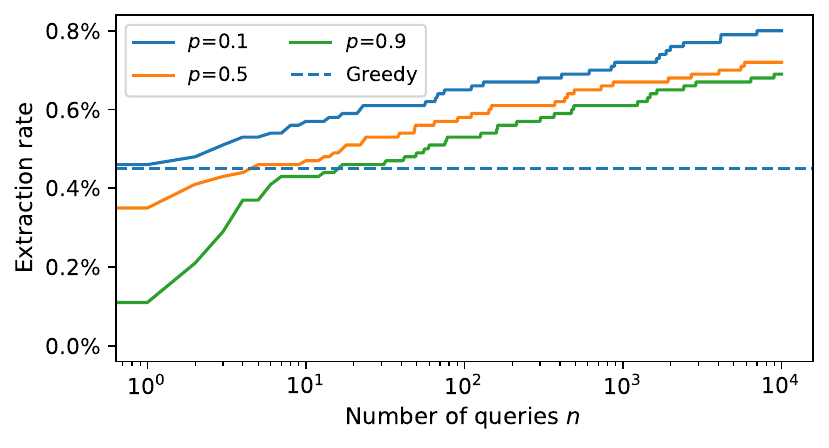}
        \caption{OPT 2.7B, top-$k$}
        \vspace{.5cm}
        \label{fig:opt-cc-2.7b-topk}
\end{subfigure}%
\hfill
\begin{subfigure}[t]{0.312\textwidth}
\centering
        \includegraphics[width=1.\linewidth]{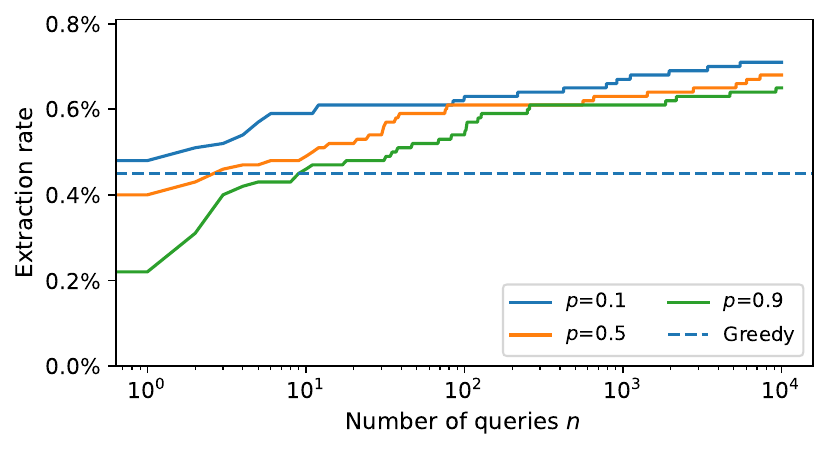}
        \caption{OPT 2.7B, top-$q$}
        \vspace{.5cm}
        \label{fig:opt-cc-2.7b-topq}
\end{subfigure}
\hfill
\begin{subfigure}[t]{0.312\textwidth}
\centering
        \includegraphics[width=1.\linewidth]{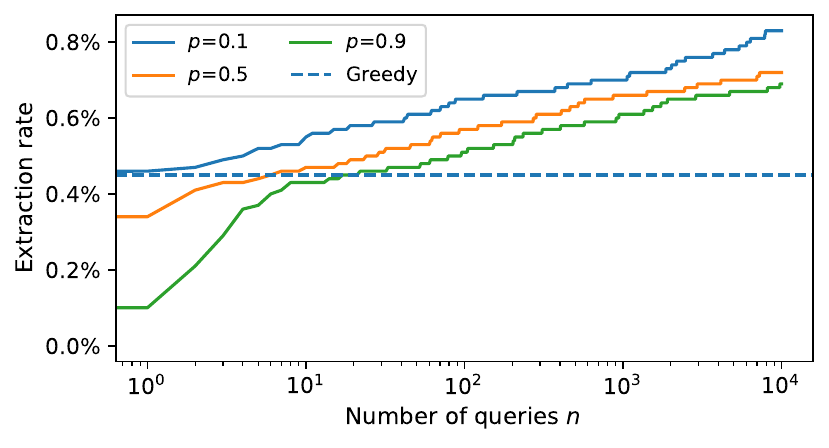}
        \caption{OPT 2.7B, $T=1$}
        \vspace{.5cm}
        \label{fig:opt-cc-2.7b-random}
\end{subfigure}\\
\begin{subfigure}[t]{0.045\textwidth}
\vspace{-1.65cm}
\textbf{6.7B}
\vspace{1cm}
\end{subfigure}%
\hfill
\begin{subfigure}[t]{0.312\textwidth}
\centering
        \includegraphics[width=1.\linewidth]{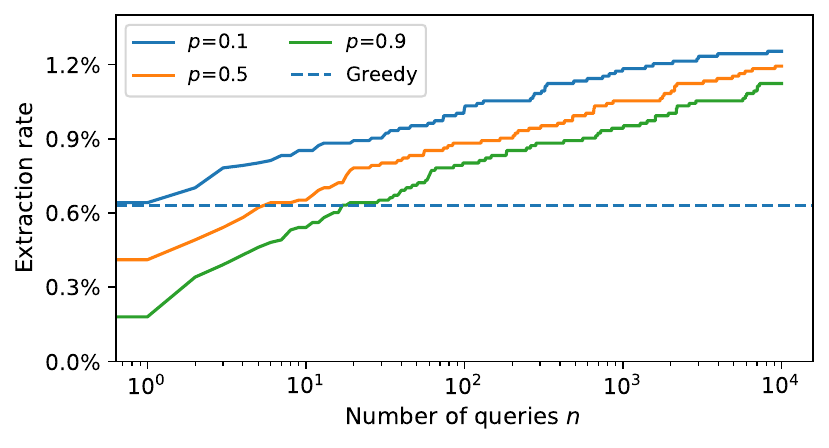}
        \caption{OPT 6.7B, top-$k$}
        \vspace{.5cm}
        \label{fig:opt-cc-6.7b-topk}
\end{subfigure}%
\hfill
\begin{subfigure}[t]{0.312\textwidth}
\centering
        \includegraphics[width=1.\linewidth]{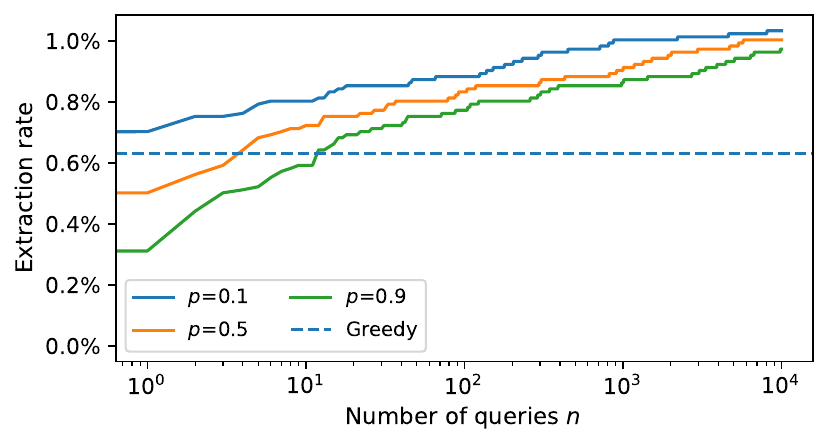}
        \caption{OPT 6.7B, top-$q$}
        \vspace{.5cm}
        \label{fig:opt-cc-6.7b-topq}
\end{subfigure}
\hfill
\begin{subfigure}[t]{0.312\textwidth}
\centering
        \includegraphics[width=1.\linewidth]{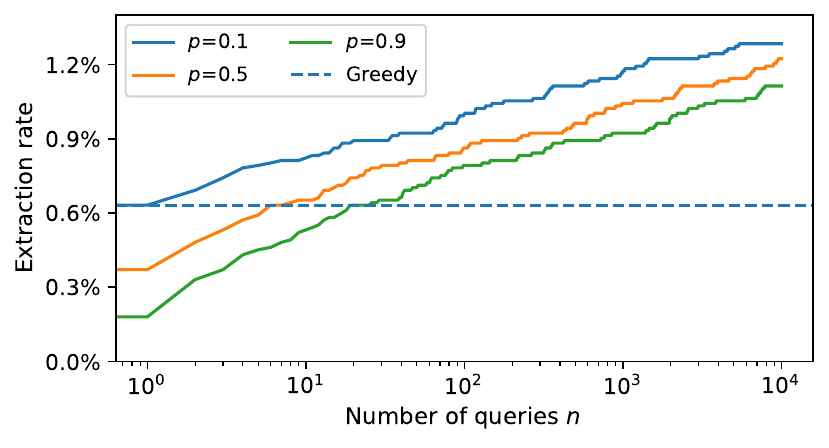}
        \caption{OPT 6.7B, $T=1$}
        \vspace{.5cm}
        \label{fig:opt-cc-6.7b-random}
\end{subfigure}\\%
\caption{\textbf{OPT models on Common Crawl.} 
Comparison of greedy-sampled discoverable extraction rates and $(n, p)$-discoverable extraction rates for different-sized OPT models (350M, 1.3B, 2.7B, and 6.7B) for different sampling schemes (top-$k$ with $k=40$ and $T=1$; top-$q$ with $q=0.1$ and $T=1$; random with $T=1$) for Common Crawl ($10,000$ examples). 
Each row is for a different model, and each column is for a different sampling scheme.
For the same model, extraction-rate curves at different values of $p$ are fairly consistent across top-$k$ and random $T=1$ sampling; top-$q$ extraction rates are slightly lower.
(This is a similar pattern as for Pythia on GitHub and Llama; see Figure~\ref{fig:pythia-github} and Figure~\ref{fig:llama-cc}, respectively.) 
The larger Llama 13B model exhibits higher extraction rates than Llama 7B. 
Larger models exhibit higher extraction rates for the plotted $n$ and $p$. 
The extraction rates for OPT 6.7B on Common Crawl are higher than for the similarly-sized Llama 7B on Common Crawl (Figure~\ref{fig:llama-cc}).
The extraction rates for OPT on Common Crawl are lower than for models of comparable sizes in the Pythia family on all data subsets (Figures~\ref{fig: app_model_sizes}, \ref{fig: sampling_schemes}, \ref{fig:pythia-wiki} \&  \ref{fig:pythia-github}).\looseness=-1
}
\label{fig:opt-cc}
\end{figure*}
\FloatBarrier

\end{document}